\documentclass{article}

\usepackage{setspace}
\RequirePackage[OT1]{fontenc}
\usepackage{verbatim}
\usepackage{multirow}
\usepackage{makecell}
\usepackage{tabularx}
\usepackage{tabu}
\usepackage{float}
\usepackage{stfloats}

\RequirePackage{amsthm,amsmath,amsfonts,amssymb,mathtools}

\usepackage[utf8]{inputenc} 
\usepackage[T1]{fontenc}    
\usepackage{url}            
\usepackage{booktabs}       
\usepackage{nicefrac}       
\usepackage{microtype}      
\usepackage{natbib}
\usepackage{doi}
\usepackage{xcolor}

\usepackage{cancel}
\usepackage{graphicx}
\usepackage{enumitem}
\usepackage{tabularx}

\usepackage{comment}

\usepackage{geometry}
\onecolumn 

\DeclareMathOperator*{\argmax}{arg\,max}
\DeclareMathOperator*{\argmin}{arg\,min}
\DeclareMathOperator{\sign}{sign}

\usepackage{hyperref} 
    \hypersetup{
    	colorlinks=true,
    	linkcolor=blue,
    	filecolor=blue,      
    	urlcolor=blue,
    	citecolor=blue
    }

\title{Artificial Intelligence-based Decision Support Systems\\ for Precision and Digital Health}
\author{Nina Deliu$^{1,2,*}$ and Bibhas Chakraborty$^{3,4,5}$\\
\normalsize $^{1}$MEMOTEF Department, Sapienza University of Rome, Italy\\ \normalsize$^{2}$MRC -- Biostatistics Unit, University of Cambridge, UK\\ \normalsize$^{3}$Centre for Quantitative Medicine and the Program in Health Services and Systems Research,\\ \normalsize Duke-NUS Medical School \\ \normalsize$^{4}$Department of Statistics and Data Science at the National University of Singapore, Singapore \\ \normalsize$^{5}$Department of Biostatistics and Bioinformatics, Duke University, USA\\ \normalsize \textit{Correspondence to}: nina.deliu@uniroma1.it}

\date{}

\begin{document}
\maketitle

\begin{abstract}
Precision health, increasingly supported by digital technologies, is a domain of research that broadens the paradigm of precision medicine, advancing everyday healthcare. This vision goes hand in hand with the groundbreaking advent of artificial intelligence (AI), which is reshaping the way we diagnose, treat, and monitor both clinical subjects and the general population. AI tools powered by machine learning have shown considerable improvements in a variety of healthcare domains. In particular, reinforcement learning (RL) holds great promise for sequential and dynamic problems such as dynamic treatment regimes and just-in-time adaptive interventions in digital health.
In this work, we discuss the opportunity offered by AI, more specifically RL, to current trends in healthcare, providing a methodological survey of RL methods in the context of precision and digital health. Focusing on the area of adaptive interventions, we expand the methodological survey with illustrative case studies that used RL in real practice.

\end{abstract}

\section{Introduction} \label{Sec: Intro}

In the current era of population aging and increased prevalence of chronic diseases, providing adequate health support remains one of the most urgent and complex global challenges~\citep{prince_burden_2015}. Chronic conditions such as cancer, diabetes, mental illness, or obesity tend to be of a long duration and often create a need for long-term treatment and care, carrying enormous social, medical, and economic burdens~\citep{beaglehole_priority_2011}. They are the result of a combination of genetic and physiological factors, as well as environmental and behavioral aspects that are challenging to modify, given the nature of modern lifestyle. Maintaining healthy behaviors throughout life, from diet and physical activity regimes to smoking habits, all contribute to reducing the risk of chronic diseases, improving physical and mental capacity, and delaying care dependency.

Precision health is a relatively nascent scientific discipline that broadens the paradigm of precision medicine by including approaches that occur outside the clinical setting~\citep{gambhir_toward_2018,ryan_defining_2021}. It seeks to develop proactive and personalized solutions to health problems, disease prevention, and health promotion. Under this framework, ``disease treatment and prevention takes into account \textit{individual variability} in genes, environment, and lifestyle for each person''~\citep[\textit{Precision Medicine Initiative};][]{collins_new_2015}. It transitions from the ``one-size-fits-all'' standards to the formulation of treatment and prevention strategies based on the unique background and condition of each patient~\citep{kosorok_precision_2019}. As an illustration of this mission, a precision health system might see health proactively co-managed by healthcare providers and patients through the synchronous integration of information, starting with genotyping at birth, regular screening, and combined continuous health monitoring and provision of actionable advice and early intervention at the precise moment when the individual needs it. 

Over the last decade, the precision health paradigm and healthcare in general have witnessed unprecedented innovation due to the continuous improvement and use of big data and digital technologies~\citep{agrawal_big_2020}. These comprise electronic tools, devices, systems, and resources that utilize increasingly fast data transmission speeds and collect, store, or process large amounts of data. Successful scientific applications of big data have already been demonstrated in numerous applications, from specific disease areas such as oncology~\citep[see e.g., \textit{The Cancer Genome Atlas} and the \textit{Pan-Cancer Analysis of Whole Genomes} initiatives;][]{tomczak_cancer_2015,aaltonen_pan-cancer_2020} or neuropsychiatry~\citep[\textit{PsychENCODE};][]{psychencode_consortium_psychencode_2015} to national initiatives. For example, the United Kingdom (UK) has now established a clear national strategy with the UK Biobank prospective cohort initiative~\citep{allen_uk_2012}, which collates together biological samples, physical measures of patient health, and sociological information such as lifestyle and demographics from 500,000 individuals.
Expanding on the UK Biobank model, the American \textit{All of US} program~\citep{all_of_us_research_program_investigators_all_2019} integrates medical records, behavioral, and family data in a unique standardized and linked database for all patients, including minorities. The goal is achieved by integrating ancillary patient data, including those collected through wearables, which are now part of daily life, interconnecting the world's population, and making health services more accessible and accountable. 

The effective use of big data in healthcare is enabled by the development and deployment of artificial intelligence (AI) approaches, such as those based on machine learning~\citep[ML;][]{bishop_pattern_2006}. ML is a subfield of AI that uses algorithms to automatically learn from past data or experiences, making it possible to unravel patterns, associations, and causations in complex and unstructured datasets created in the era of big data~\citep{camacho_next-generation_2018}. In turn, it allows one to quickly provide actionable analysis on data, generating accurate prediction models--such as response of a patient to a treatment regimen--and supporting clinical practice with increasingly better decisions. An overview of successful biomedical applications using ML is provided in \cite{deo_machine_2015} and \cite{rajkomar_machine_2019}. 

As an alternative ML area, Reinforcement Learning~\citep[RL;][]{sutton18_rl, bertsekas2019reinforcement, sugiyama2015statistical}, represents a framework for interactive tasks in which the system or algorithm must learn by interacting with the surrounding environment \textit{sequentially}. More specifically, in RL problems, at each time step of a sequential process, an \textit{agent} interacts with its \textit{environment}, performs \textit{action(s)}, and, based on a \textit{feedback} received from the environment for the selected action(s), learns, by \textit{trial-and-error}, on how to take better actions in order to maximize the cumulative feedback over time. This distinctive feature offers a powerful solution in a variety of healthcare domains where the problem has a sequential nature~\citep{chakraborty_statistical_2013, yu_reinforcement_2023,gottesman_guidelines_2019}, such as dynamic treatment regimes \citep[DTRs;][]{chakraborty_statistical_2013}. Furthermore, the continuous improvement and use of mobile technologies has determined the development of a new area for health promotion, known as mobile health~\citep[mHealth;][]{istepanian2007m}, which aims to deliver real-time interventions tailored to individual characteristics and their rapidly changing circumstances. Such interventions are termed \textit{just-in-time adaptive interventions}~\citep[JITAIs;][]{nahum2018just} and have a central position in this survey. Specifically, the focus of this work is on sequential decision-making problems in healthcare and includes DTRs and JITAIs in mHealth as two key areas that have embraced the use of AI instruments.

In the coming years, AI is expected to radically transform healthcare and the way it is delivered. AI systems supported by ML have achieved considerable improvements in accuracy for diagnosis or image-based diagnosis~\citep{myszczynska_applications_2020, mckinney_international_2020}, prognosis~\citep{kourou_machine_2015} or drug discovery~\citep{vamathevan_applications_2019}, among others. Active research in both AI and precision health points to their convergence toward a future where healthcare is enhanced with highly personalized information and healthcare providers are empowered by decision-making support systems through augmented intelligence~\citep{johnson_precision_2021}. 

Motivated by the increasing interest shown within the healthcare domain in AI technologies such as RL, this work aims to provide an overview of RL in the field of precision and digital health. Along with a survey of RL methods for specific applications in the area, we illustrate two case studies, the \textit{PROJECT QUIT - FOREVER FREE}~\citep{chakraborty_statistical_2013} and the \textit{DIAMANTE} text messaging system~\citep{aguilera_mhealth_2020,figueroa_daily_2022}, in the context of smoking cessation and physical activity, respectively, and the challenges we faced when designing the AI-based system. We believe that there is scope for important practical advances in these areas, and with this overview we aim to make it easier for methodological disciplines to join forces to assist healthcare practice and discovery and to develop the next generation of methods for AI in healthcare.

The remainder of this contribution is structured as follows. In Section~\ref{Sec: RL_framework}, we provide the mathematical formalization of the general RL framework, which is further explored in Section~\ref{Sec: MABs} with a focus on the multi-armed bandit problem. In Section~\ref{Sec: DTRs} and Section~\ref{Sec: JITAIs}, we introduce the two areas of interest, namely DTRs and JITAIs, and extensively review existing data sources and RL methodologies for these problems. Two case studies are then illustrated in Section~\ref{Sec: DTR_applications_challenges} and Section~\ref{Sec: TS_Diamante} for DTRs and JITAIs, respectively. Final considerations and concluding remarks are given in Section~\ref{Sec: Conclusions}.

\section{The Reinforcement Learning Framework} \label{Sec: RL_framework}

Reinforcement learning is an area of machine learning (ML) concerned with understanding how agents (e.g., systems or machines) might learn to improve their decisions through repeated experience. More formally, it aims to identify optimal decision rules (or \textit{policies}) in \textit{sequential decision-making problems under uncertainty}~\citep{sutton18_rl, bertsekas2019reinforcement}. An optimal RL policy is one that maximizes the expected long-term utility, assuming that this is likely to outweigh the associated short-term costs. The general RL framework is formalized through a continuous interaction between a \textit{learning agent} (i.e., the decision maker) and the \textit{environment} it wants to infer about. At each interaction stage, the agent observes some representation of the environment's \textit{state} or \textit{context}, and on that basis selects an \textit{action}, that is, makes a decision. The impact of the chosen action is evaluated through a \textit{reward} (or feedback) provided by the environment. Based on the reward received, the agent learns, by \textit{trial-and-error}, on how to take better actions in the future to maximize the cumulative reward over time.

\subsection{Basic ingredients}
In reinforcement learning, differently from other ML methods, data are characterized by a sequential order and learning is carried out through many stages. For practicality, consider a discrete time space indexed by $t \in \mathbb{N}= \{0, 1,\dots, \}$. At each time $t$, the RL framework is described as an interaction between an agent and an unknown environment, articulated in the following three key elements:
\begin{itemize}
    \item \textit{State} or \textit{context}, denoted by $X_t \in \mathcal{X}_t$, being the representation of the environment at time $t$. This includes the set of information (demographic and health-related covariates or physical data such as location) that may be relevant to understanding the consequences of alternative interventions. 
    \item \textit{Action} $A_t$, taken by the agent from a set of admissible actions $\mathcal{A}_t$, i.e., the set of alternative interventions. When making the choice $A_t$, the agent weighs the consequences of the alternatives and their likelihood, given the state $X_t$.
    \item A \textit{reward} $Y_{t+1} \in \mathcal{Y}_{t+1} \subset \mathbb{R}$ provided by the environment in response to the chosen action $A_t$ in correspondence with an observed state $X_t$. It is the information that an agent learns only after taking an action (e.g., patient response to treatment). This is closely related to the concept of utility, which should be the ultimate criterion to judge whether the entire policy works well or not. 
\end{itemize}
Once action $A_t$ is selected, together with the provision of a reward $Y_{t+1}$, the environment makes a transition to a new state $X_{t+1} \in \mathcal{X}_{t+1}$. Using this notation, the characterization of the RL sequential decision problem can be described as an ordered sequence or trajectory given by:
\begin{align}\label{eq: trajectory}
X_0 \rightarrow A_0 \rightarrow (Y_1, X_1) \rightarrow A_1 \rightarrow (Y_2, X_2) \rightarrow \dots \rightarrow A_t \rightarrow (Y_{t+1}, X_{t+1}) \rightarrow \dots
\end{align} 
In healthcare, this trajectory can be seen as the \textit{history} of the interventions received over the course of a disease or program, and the individual responses to treatment along with the time-varying contextual and individual health-related information. 

\subsection{Mathematical formalization of the general RL}

Define $\mathbf{X_t} \doteq \left(X_\tau \right)_{\tau = 0,\dots, t}$, $\mathbf{A_t} \doteq \left(A_\tau \right)_{\tau = 0,\dots, t}$, $\mathbf{Y_{t+1}} \doteq \left(Y_{\tau+1} \right)_{\tau = 0,\dots, t}$, and similarly $\mathbf{x_t}$, $\mathbf{a_t}$ and $\mathbf{y_{t+1}}$, where the upper- and lower-case letters denote random variables and their particular realizations, respectively. Also define $\mathbf{H_t}$ 
as the \textit{history} all the information available at time $t$ prior to decision $A_{t}$, i.e., $\mathbf{H_t}  \doteq (\mathbf{A_{t-1}}, \mathbf{X_t}, \mathbf{Y_t})$; similarly $\mathbf{h_t}$. The history $\mathbf{H_t}$ at stage $t$ belongs to the product set $\boldsymbol{\mathcal{H}_t} = \mathcal{X}_0 \times \prod_{\tau = 0}^{t-1}{\mathcal{A}_{\tau} \times \mathcal{X}_{\tau+1} \times \mathcal{Y}_{\tau+1}}$. Note that, by definition, $\mathbf{H_0} = X_0$. We assume that each longitudinal history is sampled independently according to a distribution $P_{\boldsymbol{\pi}}$, given by
\begin{align} \label{eq: P_pi}
P_{\boldsymbol{\pi}} \doteq p_0(x_0) \prod_{t \geq 0} \pi_{t}(a_{t} \mid \mathbf{h_{t}})
p_{t+1}(x_{t+1}, y_{t+1} \mid \mathbf{h_{t}}, a_{t}),
\end{align} 
where:
\begin{itemize}[leftmargin=*]
    \item $p_0$ is the probability distribution of the initial state $X_0$.

    \item $\boldsymbol{\pi} \doteq \{\pi_t\}_{t \geq 0}$ represents the agent's \textit{policy} and determines the sequence of actions generated throughout the decision-making process. More specifically, $\pi_t$ maps histories of length $t$, $\mathbf{h_t}$, to a probability distribution over the action space $\mathcal{A}_t$, i.e., $\pi_t( \cdot  \mid \mathbf{h_t})$. The conditioning symbol ``$ \mid $'' in $\pi_t( \cdot  \mid \mathbf{h_t})$ reminds us that the policy defines a probability distribution over $\mathcal{A}_t$ for each $\mathbf{h_t} \in \boldsymbol{\mathcal{H}_t}$. 
    Sometimes, $A_t$ is uniquely determined by the history $\mathbf{H_t}$, therefore the policy is simply a function of the form 
    $\pi_t(\mathbf{h_t}) = a_t$.  We call it \textit{deterministic policy}, in contrast with \textit{stochastic policies} that determine actions probabilistically.

    \item $\{p_t\}_{t \geq 1}$ are the unknown \textit{transition probability distributions} that characterize the dynamics of the environment. At each time $t \in \mathbb{N}$, the transition probability $p_t$ assigns to each trajectory $(\mathbf{x_{t-1}},\mathbf{a_{t-1}}, \mathbf{y_{t-1}}) = (\mathbf{h_{t-1}},a_{t-1})$ at time $t-1$ a probability measure over $\mathcal{X}_t \times \mathcal{Y}_t$, specifying the probability of transition to new states in $\mathcal{X}_t$ with reward in $\mathcal{Y}_t$, from history $\mathbf{h_{t-1}}$ and action $a_{t-1}$, i.e., $p_t(\cdot, \cdot \mid \mathbf{h_{t-1}}, a_{t-1})$. 
\end{itemize}

\begin{figure}
    \centering
    \includegraphics[scale = .7]{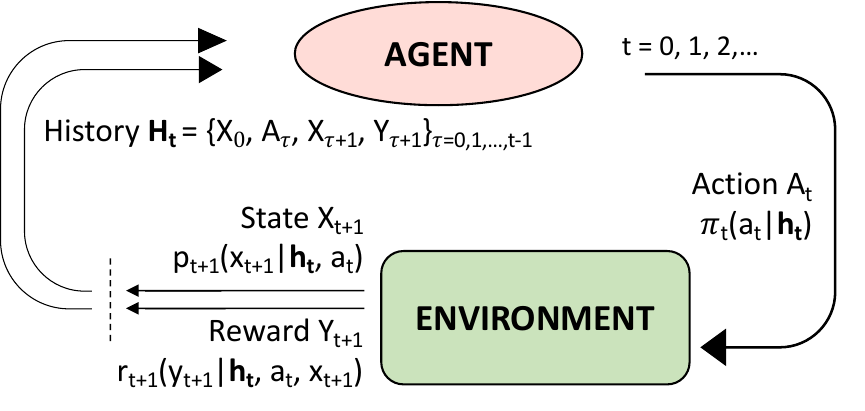}
    \caption{Illustration of the general RL framework.}
    \label{fig: RL_graph}
\end{figure}

At each time $t$, the transition probability distribution $p_{t+1}(x_{t+1}, y_{t+1} \mid \mathbf{h_{t}}, a_{t})$ gives rise to: (i) the \textit{state-transition probability distribution} $p_{t+1}(x_{t+1} \mid \mathbf{h_{t}}, a_{t})$, i.e., the probability of transitioning to state $x_{t+1}$ having observed a history $\mathbf{h_{t}}$ and taking action $a_{t}$; and (ii) the \textit{immediate reward distribution} $r_{t+1}(y_{t+1} \mid \mathbf{h_{t}},a_{t},x_{t+1})$, which specifies the reward $Y_{t+1}$ after transitioning from a history $\mathbf{h_{t}}$ to $x_{t+1}$ under action $a_t$. 
To better incorporate uncertainty, we assume a stochastic reward distribution. An illustrative representation of the RL framework is provided in Figure~\ref{fig: RL_graph}.
The cumulative discounted sum of immediate rewards from time $t$ onward is known as \textit{return}, say $\mathbf{R_t}$, and is given by
\begin{align} \label{eq: return}
\mathbf{R_t} \doteq Y_{t+1} + \gamma Y_{t+2} + \gamma^2 Y_{t+3} + \dots = \sum_{\tau \geq t} {\gamma^{\tau-t} Y_{\tau+1}}, \quad t \in \mathbb{N}.
\end{align}
The \textit{discount rate} $\gamma \in [0, 1]$ determines the current value of future rewards: a reward received $\tau$ time steps in the future is worth only $\gamma^{\tau}$ times what it would be worth if it were received immediately. 
If $\gamma = 0$, the agent is \textit{myopic} in being concerned only with maximizing the immediate reward, that is, $\mathbf{R_t} = Y_{t+1}$, with the convention that $0^0 = 1$. 
If $\gamma = 1$, the return is \textit{undiscounted} and it is well defined (finite) as long as the time-horizon is finite, i.e.,  $T < \infty$ \citep{sutton18_rl}. 
\bigskip 

\noindent\textbf{Remark: On-policy Vs. off-policy learning} The agent's policy $\boldsymbol{\pi}$ determines the sequential selection of actions. In a randomized experiment context, for example, it may define the randomization probabilities of each intervention at each decision point $t$. In such a setting, an agent can be interested in learning and optimizing the policy $\boldsymbol{\pi}$ while following it, that is, from experiences sampled directly from $\boldsymbol{\pi}$. This type of learning is termed \textit{on-policy} or \textit{online} learning, and policy $\boldsymbol{\pi}$ represents both the \textit{exploration policy} (the one that generates the data) and the \textit{target policy} (the one we learn about). In contrast, there are settings where the agent learns from previously collected data without interacting with the environment to collect samples (e.g., observational data). In this type of learning, termed \textit{off-policy} or \textit{offline}, we say that the target policy is learned from data ``off'' the target policy, which are determined according to a policy $\boldsymbol{\pi}$ that can be either the exploration policy (when known, e.g., in randomized studies), or, more generally, an observed or \textit{behavior policy}. Similar concepts are used in statistical and causal inference for referring to the estimation of unknown quantities of interest such as parameters.

Taking into account a potential misalignment between the target policy, say $\mathbf{d}$, and the one used to generate the data $\boldsymbol{\pi}$ (either exploration or behavior), the RL problem at any time $t$ is to learn an optimal way to choose the set of actions, i.e., an \textit{optimal policy} $\mathbf{d^*_t} \doteq \{d^*_t\}_{\tau \geq t}$, so as to maximize the expected future return. Formally,
\begin{align}\label{eq: opt_policy}
    \boldsymbol{d_t^*} = \argmax_{\boldsymbol{d_t}}\mathbb{E}_{\boldsymbol{d}}[\mathbf{R_t}] = \argmax_{\boldsymbol{d_t}} \mathbb{E}_{\boldsymbol{d}} \left[ \sum_{\tau \geq t} {\gamma^{\tau-t} Y_{\tau+1}} \right],
\end{align}
where the expectation is meant with respect to a trajectory distribution $P_\mathbf{d}$ analogous to Eq.~\eqref{eq: P_pi}, where the policy $\boldsymbol{\pi}$ that generated the data is replaced by the target policy $\mathbf{d}$ we want to learn about. Note that the \textit{expected} return is the most common approach to handle decision making under uncertainty~\citep{de_lara_sustainable_2008}. 

For learning optimal policies, various methods have been developed so far in the RL literature: see \cite{sutton18_rl} and \cite{sugiyama2015statistical} for an overview. A traditional approach is through \textit{value functions}, which define a partial ordering over policies, with insightful information on the optimal ones. In fact, optimal policies share the same (optimal) value function, and comparing estimated value functions of different candidate policies offers a way to understand which strategy may offer the greatest expected outcome.

There are two types of value functions: i) \textit{state-value} or simply \textit{value} functions, say $V_t^{\boldsymbol{d}}$, representing how good it is for an agent to be in a given state, and ii) \textit{action-value} functions, say $Q_t^{\boldsymbol{d}}$, indicating how good it is for the agent to perform a given action in a given state.
These are formally defined as: 
\begin{align} 
    V_t^{\boldsymbol{d}} (\mathbf{h_t}) &\doteq
    \mathbb{E}_{\boldsymbol{d}}\left[ \mathbf{R_t} | \mathbf{H_t} = \mathbf{h_t} \right] = 
    \mathbb{E}_{\boldsymbol{d}}\left[ 
    \sum_{\tau \geq t} {\gamma^{\tau-t} Y_{\tau+1}} \middle| \mathbf{H_t} = \mathbf{h_t} 
    \right], \label{eq: value} \\
    Q_t^{\boldsymbol{d}} (\mathbf{h_t}, a_t) &\doteq 
    \mathbb{E}_{\boldsymbol{d}}\left[ \mathbf{R_t} | \mathbf{H_t} = \mathbf{h_t}, A_t = a_t \right] = 
    \mathbb{E}_{\boldsymbol{d}}\left[ 
    \sum_{\tau \geq t} {\gamma^{\tau-t} Y_{\tau+1}} \middle| \mathbf{H_t} = \mathbf{h_t}, A_t = a_t \right], \label{eq: q-value}
\end{align}
$\forall t \in \mathbb{N}$, $\forall \mathbf{h_t} \in \boldsymbol{\mathcal{H}_t}$ and $\forall a_t \in \mathcal{A}_t$, with $\mathbf{H_t}$ and $A_t$ such that $\mathbb{P}(\mathbf{H_t} = \mathbf{h_t}) > 0$ and $\mathbb{P}(A_t = a_t) > 0$. 
By definition, at stage $t=0$, $V_0^{\boldsymbol{\pi}}(\mathbf{h_0})\doteq V_0^{\boldsymbol{\pi}}(x_0)$; while for the terminal stage, if any, the state-value function is $0$.

At stage $t$, the \textit{optimal value function} $V_t^{\boldsymbol{d^*}}$ yields the largest expected return for each history, and the \textit{optimal Q-function} $Q_t^{\boldsymbol{d^*}}$ yields the largest expected return for each history-action pair, i.e.,
\begin{align} \label{eq: optQ}
    Q_t^*(\mathbf{h_t}, a_t) &\doteq \max_{\boldsymbol{d_t}}Q_t^{\boldsymbol{d}}(\mathbf{h_t}, a_t), \quad \forall \mathbf{h_t} \in \boldsymbol{\mathcal{H}_t}, \forall a_t \in \mathcal{A}_t.
\end{align}

A fundamental property of the value functions used throughout RL is that they satisfy particular recursive relationships, known as \textit{Bellman} equations~\citep{Bellman1957}. In the Q-value case, for instance, for any policy $\mathbf{d}$, the following consistency condition, expressing the relationship between the quality of an history-action and the quality of the successors, holds:
\begin{align} \label{eq: bellman_opt_Q}
   Q_t^*(\mathbf{h_t}, a_t) = \mathbb{E} \Big[ & Y_{t+1} + \gamma \max_{a_{t+1} \in \mathcal{A}_{t+1}}Q_{t+1}^*(\mathbf{h_{t+1}}, a_{t+1}) \mid  \mathbf{H_t} = \mathbf{h_t}, A_t = a_t \Big], 
\end{align}
$\forall a_t \in \mathcal{A}_t, \forall \mathbf{h_t} \in \boldsymbol{\mathcal{H}_t}, \forall t \in \mathbb{N}$, and with discrete state and action spaces. Here, the expectation $\mathbb{E}$ is taken with respect to the transition distribution $p_{t+1}$ only, which does not depend on the policy; thus, the subscript $\boldsymbol{d}$ can be omitted. 

The property in Eq.~\eqref{eq: bellman_opt_Q} allows estimation of (optimal) value functions recursively, from $T$ backward in time. In finite-horizon \textit{dynamic programming} (DP), this technique is known as \textit{backward induction} and represents one of the main methods to solve the Bellman equation. In infinite- and indefinite-horizon problems, traditional backward induction is not possible, given the impossibility of extrapolating beyond the time horizon in the observed data. To overcome this issue, alternative methods and additional assumptions (e.g., discounting and boundedness of rewards) are typically taken into account. Common strategies~\citep[e.g., V-learning, which we review in Section~\ref{Sec: DTRs_direct};][]{luckett2020estimating}, focuses on time-homogeneous Markov processes.

Due to its generality, RL is studied and employed in many disciplines, from game theory to education and healthcare. In this work, our goal is to review common RL classes that have been studied or used to support decision making in precision and digital health. In particular, we cover the RL methods studied in the DTR literature (e.g., Q-learning and outcome weighted learning) and in digital health (with a main interest in the multi-armed bandit class, which we discuss in Section~\ref{Sec: MABs}). For readers interested in the general area of RL and the broad spectrum of existing methods, without a specific application in mind, we refer to~\cite{sutton18_rl, sugiyama2015statistical}.

\section{Dynamic Treatment Regimes in Precision Health} \label{Sec: DTRs}

Clinical or behavioral treatments often involve a series of decisions over time that account for the continuously evolving histories of individuals. For example, weight loss management involves a sequence of decisions at multiple stages of weight progression. Initially, individuals affected by excess body weight undergo lifestyle modifications (such as diet and exercise), and based on their body mass index (BMI), may be treated with pharmacologic therapies to achieve the desired body weight. Then, if the individual \textit{responds} (i.e., shows a significant weight loss), the physician may prescribe a maintenance therapy (typically diet and exercise) to maintain weight at a reduced level. Otherwise, the clinician prescribes a \textit{second-line} therapy, to try to induce body weight reduction. There exist many possible therapies. The aim of the physician is to choose the sequence of therapies that leads to the best possible outcome, e.g., long-term maintenance of lost weight, for that individual. 

Similarly, the treatment of cancer, diabetes, mental health disorders, or the management of addiction problems requires a series of decisions by which the physician can start, stop, maintain, modify, or adjust interventions based on the patient's response and evolving characteristics. This sequence of decisions constitutes a \textit{dynamic treatment regime} or \textit{regimen}~\citep{murphy_optimal_2003,chakraborty_statistical_2013}, alternatively known as adaptive interventions or strategies~\citep{collins_conceptual_2004,lavori_design_2000}.

Dynamic treatment regimes (DTRs) offer a vehicle to operationalize the sequential decision-making process involved in clinical practice and can also be viewed as a \textit{decision support system}. A DTR is defined as a sequence of decision rules, one per stage of intervention, dictating how to personalize treatments to patients based on their baseline and evolving history (\textit{time-varying}, \textit{dynamic state}), repeatedly adjusting over time in response to ongoing performance~\citep{almirall_introduction_2014,nahum2018just}. 
Thus, the treatment regime is ``dynamic'' within a person over time, varying because the person or disease is changing, with the goal of obtaining the best results for that individual. 

The existing DTR frameworks~\citep{collins_conceptual_2004,almirall_introduction_2014} highlight four components that play an important role in the design of these interventions: 
\begin{enumerate}[label = (\roman*)]
    \item The critical \textbf{decision points}, specifying the time points at which a decision concerning intervention (e.g., continue, alter, add, or subtract treatment) has to be made; here we assume a finite or countable number of times $t = 0, 1,\dots$;
    \item The \textbf{decisions or treatment options} at each time $t$,  denoted by $A_t \in \mathcal{A}_t$, where $\mathcal{A}_t$ is the decision or action space, generally discrete;
    \item The \textbf{tailoring variable(s)} at each time $t$, say $X_t \in \mathcal{X}_t$, with $\mathcal{X}_t \subseteq \mathbb{R}^p$, capturing individuals' baseline and time-varying information for personalizing decision-making;
    \item The \textbf{decision rules} $\mathbf{d} = \{d_t\}_{t \geq 0}$, that, at each time $t$, link the tailoring variable(s) to specific decisions.
\end{enumerate}

Treatment options $A_t \in \mathcal{A}_t$ are not limited to different medications or drugs, but can also include different dosages~\citep[duration, frequency or amount; ][]{voils_informing_2012,chen_personalized_2016}, various tactical options (for example, increase, change, maintain), modes of administration (for example, oral or injection), timing schedules~\citep{nie_learning_2021}, behavioral interventions, or no further treatment. Tailoring variables $X_t \in \mathcal{X}_t$ refer to patient and treatment information available up to the time of the critical decision, and may include previous treatment and disease history, genetic information, diagnostic test results, etc. Once the four elements are defined, each decision rule $\mathbf{d} = \{d_t\}_{t \geq 0}$ takes the individual characteristics $X_t \in \mathcal{X}_t$ of a subject and their treatment history observed up to that stage $\{A_t\}_{t = 0, 1, \dots, t-1}$ as input and outputs a recommended treatment strategy at that stage. The dynamic treatment regime $\mathbf{d_t} = (d_0,\dots,d_t)$ is regarded a multistage regime with each $d_\tau$, $\tau = 0,\dots,t$ being a mapping of the entire evolving history $\mathcal{X}_0\times\mathcal{A}_0 \times \dots \times \mathcal{A}_{\tau-1} \times \mathcal{X}_\tau$ to $\mathcal{A}_\tau$.  Unlike average-based single-stage protocols, DTRs explicitly incorporate the heterogeneity in treatment effect among individuals and \textit{across time} within an individual. As such, it provides an attractive framework for personalized treatments in longitudinal settings. Furthermore, by treating only those who show a need for treatment, DTRs hold the promise of reducing noncompliance due to overtreatment or undertreatment~\citep{lavori_design_2000}.

\subsection{RL methods for constructing optimal DTRs} \label{Sec: DTRs_methods}

One of the main research goals in the field of personalized dynamic treatments is to construct \textit{optimal} DTRs, that is, to identify the treatment rule(s) that result in the best (typically long-term) mean outcome, i.e., with the highest utility. Most attempts to achieve this goal essentially require knowing or estimating the prespecified \textit{utility} function or some variations of it. For example, \cite{murphy_optimal_2003} defines \textit{regret} (i.e., \textit{loss}) functions, while \cite{robins2004optimal} introduces \textit{blip} functions~\citep{kitagawa_who_2018}, alternatively known as \textit{welfare gains} in econometrics. 

Methodologies for estimating optimal DTRs are of considerable interest within the domain of precision health and comprise a growing body of research in both computer science and statistics~\citep{chakraborty_statistical_2013}. On the one hand, the sequential decision-making nature of DTR problems perfectly conforms to the RL framework, thus attracting increasing attention in the ML literature. On the other hand, the need to quantify causal relationships, rather than mere associations, called for the intervention of the causal inference community. Since the underlying system dynamics is often unknown, inferring the consequences of executing a policy $\mathbf{d} = \{d_t\}_{t \geq 1}$ and understanding the causal effects on an outcome is a challenging task. We refer to \cite{deliu_dynamic_2022} and \cite{tsiatis_dynamic_2021} for the broad range of aspects related to DTR (including inference), while here we focus exclusively on the role of AI, more specifically RL methods, in deriving optimal DTRs. Notably, due to the similarity between the two problems and their components, RL represents one of the main approaches employed in the DTR literature. A preliminary non-exhaustive correspondence table between the RL and DTR terminologies is reported in \textbf{Table~\ref{tab: terminology}}.
\begin{table}[htb]
\centering
    \caption{Terminology correspondence between reinforcement learning (RL) and dynamic treatment regimes (DTRs).}
    \begin{tabularx}{\textwidth}{l|l|X}
    \hline
        \multicolumn{1}{c|}{\textbf{Notation}} &\multicolumn{2}{c}{\textbf{Terminology}}\\
        & RL & DTRs \\
        \hline 
        $i$& Trajectory, Unit & Patient, Subject, Individual\\
        $t$& Time, Step, Round & Stage, Interval, Round\\
        $X$& State, Context& Covariates\\
        $A$& Action, Arm & Treatment, Intervention\\
        $Y$& Reward, Feedback & Outcome, Response\\
        $\mathbf{H}$ & History, Filtration & Time Varying History\\
        $\boldsymbol{\pi}$/$\mathbf{d}$& Policy & Dynamic Treatment Regime(n), Adaptive Intervention\\
        \hline
    \end{tabularx}
    \label{tab: terminology}
\end{table}

Most of the existing work in DTRs rely on the finite-horizon setting (with a prespecified horizon $T < \infty$), and the strongly connected \textit{offline learning} procedures, wherein estimation is based on existing longitudinal data. Recent solutions in the indefinite-horizon setting, particularly suitable for electronic health record (EHR) data and for chronic diseases--where the number of stages can be arbitrarily large and are not known a-priori--are proposed in \cite{ertefaie2018constructing} and \cite{luckett2020estimating}. 

\begin{figure}[htb]
    \centering
    \includegraphics[width=.9\linewidth]{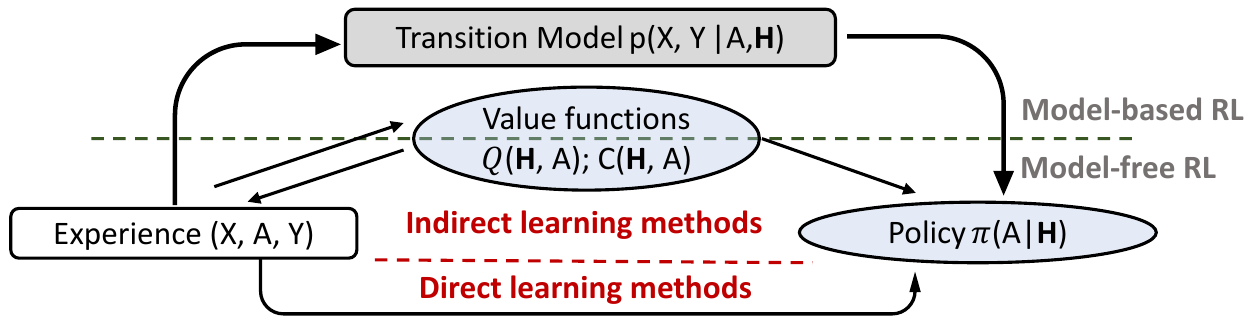}
    \caption{Illustrative comparison of direct vs indirect RL methods.}
    \label{fig: RL-taxonomy}
\end{figure}

The fundamental learning mechanism for deriving optimal policies consists in two approaches: \textit{direct} and \textit{indirect} methods. Direct methods learn optimal policies by directly looking for the policy that maximizes an objective (typically the expected return or value function) within a class of policies. On the contrary, indirect methods attempt first to estimate a value function and then to determine an optimal policy based on the learned value function. In the RL literature, direct and indirect methods are sometimes referred to as \textit{model-free} and \textit{model-based} algorithms~\citep{sutton18_rl}. However, more subtle classifications~\citep[see e.g.,][]{guan2019direct, sugiyama2015statistical} tend to make a clearer separation between the two categories in the sense that direct/indirect are used for the learning process, while model-free/model-based refer to the modeling assumptions for the environment. A graphical illustration is provided in Figure~\ref{fig: RL-taxonomy}.
In what follows, we review existing RL techniques for developing DTRs, covering both finite- and indefinite-horizon settings, and adopting the direct vs. indirect taxonomy, in line with the current DTR literature~\citep{chakraborty_dynamic_2014, deliu_dynamic_2022}. 
\bigskip

\noindent \textbf{Remark: Causal inference} For simplicity of exposition, the illustration of this subject will be carried out in a simplified framework where the main assumptions of causal inference~\citep[see e.g.,][]{chakraborty_dynamic_2014} hold. It follows that RL can operate in a simplified causal inference problem in which actions are unconfounded. For a comprehensive treatment of causal inference, we refer to~\cite{hernan_causal_2023}, while for a specific characterization of the framework for the DTR problem, we refer to~\cite{chakraborty_dynamic_2014, deliu_dynamic_2022}.

\subsection{Indirect RL methods in DTRs} \label{Sec: DTRs_indirect}
Indirect methods focus on estimating an optimal objective function (typically, an expectation of the outcome variable such as the Q-function presented in Eq.~\eqref{eq: q-value}), and then obtaining the associated policy. These methods are mainly based on iterative techniques such as dynamic programming (DP) and approximate dynamic programming (ADP), and include the Q-learning~\citep{murphy2005generalization} approach that we illustrate below. 

We mainly focus on the finite-horizon setting, where the utility function is optimized over a fixed and prespecified period of time $T$, and, for the sake of simplicity, we consider deterministic policies which map histories $\mathbf{h}$ directly into actions or decisions, that is, $\mathbf{d}(\mathbf{h}) = \mathbf{a}$.
\bigskip

\noindent \textbf{Q-learning} 
Q-learning~\citep{watkins89_rl} represents the core of modern RL and one of the most popular strategies in DTR research. Its fundamental idea is based on iterative improvement of the estimates of the Q-function at a given stage $t$, starting from a previous estimate and following the Bellman rule in Eq.~\eqref{eq: bellman_opt_Q}. That is,
\begin{align*}
    Q_t(\mathbf{h_t}, a_t) &\leftarrow Q_t(\mathbf{h_t}, a_t) + \alpha_t \Big[Y_{t+1} +\gamma \max_{a_{t+1} \in \mathcal{A}_{t+1}} Q_{t+1}(\mathbf{h_{t+1}}, a_{t+1}) - Q_t(\mathbf{h_t}, a_t) \Big].
\end{align*}
The constant $\alpha_t$ determines to what extent the newly acquired information should override the old information, i.e., how fast learning takes place: a factor of $0$ will make the learner not learn anything, while a factor of $1$ would make the learner fully update based on the most recent information. The discount factor $\gamma$ balances the immediate rewards of the learner with future rewards, and in a finite-horizon problem it is generally set to one.

The original version of this approach is known as tabular Q-learning, and it is based on storing the Q-function values for each possible state and action in a lookup table and choosing the one with the highest value. Under some appropriate and rigorous assumptions~\citep{watkins89_rl}, $Q_t$ has been shown to converge to the optimal Q-function $Q_t^*$ with probability $1$. However, this procedure is practical for a small number of problems because it can require many thousands of training iterations to converge. In addition, it represents value functions in arrays or tables, based on each state and action. Thus, large state spaces lead not just to memory issues for large tables but also to time problems needed to fill them accurately. A more recent version of Q-learning, known as \textit{Q-learning with function approximation} (FA), offers a powerful and scalable tool to overcome both the modeling requirements and the computational burden to solve an RL problem through backward induction. 

The main idea of Q-learning with FA is first to estimate the Q-functions using an approximator, e.g., regression models, neural networks or decision trees, and then to derive the estimated policy based on the estimated Q-functions. Considering an approximation space for each of the $T$ stage-specific Q-functions, e.g., $\mathcal{Q}_t \doteq \left \{Q_t(\mathbf{h_t}, a_t; \theta_t): \theta_t \in \Theta_t \right \}$, with $\Theta_t$ the parameter space, an optimal stage-$t$ policy estimate is given by:
\begin{align*}
    \hat{d}_t^*(\mathbf{h_t}) &= \argmax_{a_t \in \mathcal{A}_t} \hat{Q}_t^*(\mathbf{h_t}, a_t) \doteq \argmax_{a_t \in \mathcal{A}_t} Q_t^*(\mathbf{h_t}, a_t; \hat{\theta}_t) \doteq d_t^*(\mathbf{h_t}; \hat{\theta}_t)),
    \quad t = 0,\dots,T.
\end{align*}

An optimal regime $\hat{\mathbf{d}}^* = (d_1^*(x_1; \hat{\theta}_1), d_2^*(\mathbf{h_2}; \hat{\theta}_2), \dots, d_T^*(\mathbf{h_T}; \hat{\theta}_T))$ is obtained by following Bellman's optimality equation in Eq.~\eqref{eq: bellman_opt_Q}, and by recursively estimating $Q^*_t$ backward in time $t = T, T-1, \dots, 1$. Noticing that Q-functions are conditional expectations, regression models represent a natural approach. For the complete general iterative procedure, as well as more specific examples, we point to \cite{deliu_dynamic_2022}. By using generalized linear models, one may extend the Q-learning method to binary and count
outcomes, and an accelerated failure time model can be incorporated for survival outcomes. In the DTR arena, Q-learning generalizations to diverse outcomes have been implemented for censored data~\cite{goldberg_q-learning_2012,zhao_reinforcement_2011,zhao_constructing_2020}, binary data~\cite{moodie_q-learning_2014}, or composite measures attempting to balance different objectives~\citep{laber_set-valued_2014}, among others.

In order for $\mathbf{\hat{d}}^*$ to be a consistent estimator of the true optimal regime $\mathbf{d}^*$, it is important to recognize that all the models for the Q-functions should be correctly specified~\citep{schulte2014q}. To address this problem, several FA alternatives such as \textit{support vector regression} and \textit{extremely randomized trees}~\citep{zhao_reinforcement_2009}, or \textit{deep neural networks}~\citep{liu2017deep,atan2018deep, raghu_continuous_2017} have been proposed. We now illustrate the latter, given the attention it has attracted in recent years.
\bigskip

\begin{figure}[htb]
    \centering
    \includegraphics[width=.8\linewidth]{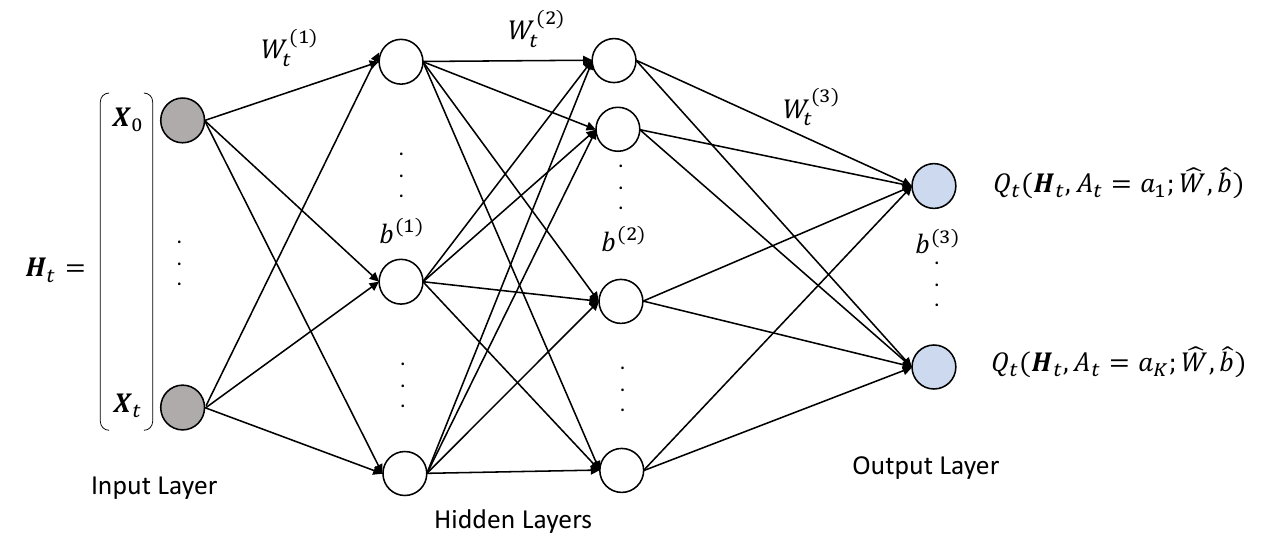}
    \caption{Representation of a feed-forward neural network with four layers used within Q-learning.}
    \label{fig: ff-nn}
\end{figure}

\noindent \textbf{Deep Q-learning.} The success achieved by Q-learning in many complex domains has been largely enabled by the use of advanced FA techniques such as \textit{deep neural networks}~\citep{mnih2015human}. We call this approach \textit{Deep Q-learning} (DQL). In DQL, a neural network~\citep{goodfellow2016deep} is used to approximate the Q-function. More specifically, at each time $t$, a DNN is used to fit a model for the Q-function in a supervised way. States and actions $\{(\mathbf{H}_{t,i}, A_{t,i})\}_{i=1,\dots,N}$ are given as inputs (in the input layer), and the Q-values of all possible actions are generated as outputs $\{Q_t(\mathbf{H}_{t,i}, A_{t,i}; \hat{\mathbf{W}},\hat{\mathbf{b}})\}_{i=1,\dots,N}$ (in the output layer), leading to a labeled set of data $ \{ (\mathbf{H}_{t,i}, A_{t,i}), Q_t(\mathbf{H}_{t,i}, A_{t,i}; \hat{\mathbf{W}},\hat{\mathbf{b}})\}_{i=1,\dots,N}$. Input data are non-linearly transformed based on the unknown weight $W$ and bias $b$ parameters and carried out through the neurons of the hidden layers. Figure~\ref{fig: ff-nn} shows a schematic of a \textit{feed-forward neural network} used within RL. It is characterized by a set of neurons, structured in layers, where each neuron processes the information from one layer to the next.
The collected data are stored and used for continuously updating the Q-function parameter estimates. To allow exploration, each decision is determined by an exploration scheme (typically $\epsilon$-greedy) that probabilistically chooses between the action with the highest Q-value and a random action. 

Within the DTR literature, DQL has been implemented with EHR data in~\cite{liu2017deep} and \cite{raghu_continuous_2017} for graft-versus-host disease and sepsis treatment, respectively. Compared to its shallow counterpart, the DQL framework is particularly suitable for (i) automatically extracting and organizing discriminative information from the data and (ii) exploring the high-dimensional action and state spaces and making personalized treatment recommendations. 
\bigskip

\subsection{Direct RL methods in DTRs} \label{Sec: DTRs_direct}

Direct methods, also known in the RL literature as direct policy search methods~\citep{ng2000algorithms}, seek to maximize the return by learning the optimal policy directly, without involving the estimation of intermediate quantities such as optimal Q-functions or contrasts. These methods typically do not assume models for conditional mean outcomes; thus, they are referred to as ``nonparametric''. However, they may consider a parameterization for the policies or regimes class.

In direct methods, a class of policies $\mathcal{D}$, often indexed by a parameter, say $\psi \in \Psi$, is first prespecified. Then, for each candidate regime $d \in \mathcal{D}$, an estimate of the corresponding \textit{utility} is obtained. The utility can be a summary of one outcome, such as percent days of abstinence in an alcohol dependence study or a composite outcome. For example, in \cite{wang2012evaluation} the utility is a compound score that combines information on treatment efficacy, toxicity, and risk of disease progression. Here, without loss of generality, we take the utility to be the value of the policy; see Eq.~\eqref{eq: value}. The regime in $\mathcal{D}$ that maximizes the value function is the estimated optimal DTR, that is, $\hat{d}^* \doteq \argmax_{d \in \mathcal{D}}\hat{V}_d$, or $\hat{d}^*  \doteq \argmax_{\psi \in \Psi}\hat{V}_{d_\psi}$ for parametric classes.
A common example of parametric classes is the \textit{soft-max} class $\mathcal{D} \doteq \{\pi(a_k| \mathbf{x}, \mathbf{\psi}) = e^{-\mathbf{x}^T\psi_k}/\sum_{j=1}^{K}{e^{-\mathbf{x}^T\psi_j}}: \mathbf{\psi} \in \Psi, k=1,\dots,K \}$, where $a_1, \dots, a_K$ are the $K$ possible treatments and $\psi \doteq (\psi^T_1,\dots, \psi^T_{K})$ is the vector of parameters for the $K$ treatments indexing the class of policies.

Most statistical work in this area is based on the \textit{inverse probability of treatment weighting} (IPTW) estimator~\citep{robins1994correcting}, used for estimating value functions~\citep{zhang2012robust, zhang2013robust}, in classification-based frameworks such as \textit{outcome weighted learning}~\citep[OWL;][]{zhao2012estimating, zhao2015new, liu2018augmented}, and in combination with ML approaches such as decision trees~\citep{laber_tree-based_2015, tao2018tree}.
Particularly useful in observational data, where the exploration and target policies differ, IPTW~\citep{robins2000marginal} 
makes use of importance sampling to change the distribution under which the regime's value is computed. In doing so, assuming that $P_d$ is absolutely continuous with respect to $P_\pi$, it basically weights outcomes according to the relative probability of interventions occurring under the target $\mathbf{d}$ and exploration $\boldsymbol{\pi}$ policies. The value function then can be rewritten as:
\begin{align}
    V^d &= \mathbb{E}_d[Y] = \int{YdP_d}= \int{Y \left( \frac{dP_d}{dP_\pi} \right)dP_\pi}=\\ &= \int \left(\prod_{t=0}^T {\frac{\mathbb{I}[A_t = d_t(H_t)]}{\pi_t(A_t|H_t)}} \right)YdP_\pi \doteq \int w_{d,\pi}YdP_\pi. \nonumber
\end{align}

To estimate $V^d$, the Monte Carlo (MC) estimator given by $\hat{V}^d \doteq \mathbb{P}_N\left [w_{d,\pi}Y\right]$, where $\mathbb{P}_N$ denotes the empirical average over $N$ trajectories, is generally used. By the Strong Low of Large Numbers, the MC estimator is unbiased, but its variance is unbounded. To stabilize this estimator, the weights $w_{d, \pi}$ are normalized by their sample mean, leading to the IPTW estimator:
\begin{align} \label{eq: IPTW_robins2000}
    \hat{V}^{d}_{IPTW} \doteq \frac{\mathbb{P}_N\left [w_{d,\pi}Y\right]}{\mathbb{P}_N\left [w_{d,\pi}\right]}.
\end{align}

The technique also allows balancing the confounders across levels of treatment: the higher the probability of receiving a specific treatment conditioned on the confounder $X$, $\pi(A|X)$, the lower the weight $w_\pi = 1/\pi(A|X)$ of their outcome $Y$. 

When $\pi$ is known (e.g., randomized trials), the IPTW estimator is consistent, but it can be highly variable due to the presence of nonsmooth indicator functions inside the weights. An alternative version, which integrates the properties of the IPTW estimator with those of regression--assuming models for both the propensity score and the (conditional) mean outcome--is the \textit{augmented inverse probability of treatment weighting} (AIPTW) estimator~\citep{zhang2012robust}. Assume a single-stage treatment regime with two treatment options ($A \in \{a, a'\}$), and let $H = X_0$ to be a patient's history, $d(H) \doteq d(H;\psi)$ a treatment regime indexed by $\psi$, $\mu(A, H;{\hat{\beta}})$ an estimated model for the mean outcome $\mathbb{E}[Y|H, A]$, and $\pi(A|H,{\hat{\gamma}})$ an estimated propensity score. Then, the AIPTW estimator is defined as:
\begin{align*}
    \hat{V}^d_{AIPW} \doteq \mathbb{P}_N \Bigg\{ &\frac{\mathbb{I}[A = d(H;\psi)]Y}{\pi(H; {\psi, \hat{\gamma}})} - 
    \frac{\mathbb{I}[A = d(H;\psi)] - \pi(H;{\psi, \hat{\gamma}})}{\pi(H;{\psi, \hat{\gamma}})}\times \mu(H;{\psi, \hat{\beta}}) \Bigg\},
\end{align*}
where,
\begin{align*}
   \pi(H;{\psi, \hat{\gamma}}) &\doteq \pi(a|H,{\hat{\gamma}})\mathbb{I}[d(H;\psi) = a] + \pi(a'|H,{\hat{\gamma}})\mathbb{I}[d(H;\psi) = a'],\\
   \mu(H;{\psi, \hat{\beta}}) &\doteq \mu(a, H;{\hat{\beta}})\mathbb{I}[d(H;\psi) = a] + \mu(a', H;{\hat{\beta}})\mathbb{I}[d(H;\psi) = a'].
\end{align*}
It only requires that either the propensity or mean outcome model to be correctly specified but not both; hence, the \textit{doubly robust} property. In addition to being more robust to model misspecification, AIPW estimators tend to be more efficient than their nonaugmented counterparts~\citep{robins2004optimal}.

Although its original version was designed for a single-stage treatment regime, it was subsequently adapted to two or more decision points~\citep{zhang2013robust, tao_adaptive_2017,zhou2018personalizing}, where models are posited for either Q-functions or contrasts. 
\bigskip

\noindent\textbf{Outcome weighted learning (OWL)} As an alternative direct approach, \cite{zhao2012estimating} studied the DTR estimation problem as a weighted classification problem--with weights retrospectively determined from clinical outcomes (hence ``Outcome Weighted'')--and proposed to solve it with ML tools (hence ``Learning'').   

In the case of two treatments, expressed as $A \in \{-1, 1\}$, \cite{qian_performance_2011} first showed that the problem can be formulated as a weighted $0-1$ loss in a weighted binary classification problem, where $d^*$ can be estimated as:
\begin{align*}
    \hat{d}^* \doteq \argmax_{d \in \mathcal{D}} \hat{V}^d &=  \argmax_{d \in \mathcal{D}} \mathbb{P}_N\left[
    \frac{\mathbb{I}[A = d(H)]}{\pi(A|H)}Y
    \right] = \argmin_{d \in \mathcal{D}} \mathbb{P}_N\left[
    \frac{\mathbb{I}[A \neq d(H)]}{\pi(A|H)}Y
    \right].
\end{align*} 

However, due to the discontinuous indicator function, \cite{zhao2012estimating} proposed to address the optimization problem with a convex surrogate loss function for the $0-1$ loss, corresponding to the \textit{hinge loss} in ML~\citep{hastie_elements_2009}. Considering that $d(H)$ can be represented in terms of the sign function $\sign(f(H))$ for some suitable function $f$, the minimization problem is then expressed as:
\begin{align}\label{eq: owl}
    \hat{f}^* \doteq \argmin_{f \in \mathcal{F}} \mathbb{P}_N\left[
    \frac{Y}{\pi(A|H)}\phi(A f(H)) + \lambda_N ||f(H)||^2
    \right],
\end{align}
where $\lambda_N$ is a tuning penalty parameter that penalizes the complexity of functions $f$, $\phi(x) \doteq \max(1-x, 0)$ is the hinge loss, and $||\cdot||$ is the norm function.

An extensive literature has considered some kind of extensions of the standard OWL estimator in Eq.~\eqref{eq: owl}, and we outline some of these in Table~\ref{tab: OWL_extensions}. In particular, \cite{zhao2015new} and \cite{liu2018augmented} have extended the OWL estimator to a multi-stage setting, proposing the \textit{Backward Outcome Weighted Learning} (BOWL) and \textit{Simultaneous Outcome Weighted Learning} (SOWL) procedures. In the first approach, the stage-$t$ estimator, denoted by $\hat{f}_{B,t}^{*}$, is obtained recursively as:
\begin{align*}
    \hat{f}_{B,t}^{*} \doteq \argmin_{f \in \mathcal{F}} \mathbb{P}_N\Bigg[
    &\frac{Y \prod_{\tau = t+1}^T\mathbb{I}[A_\tau = \hat{d}^*_\tau(\mathbf{H}_\tau)]}{ \prod_{\tau=t}^T\pi_\tau(A_\tau|\mathbf{H}_\tau)}\phi(A_t f_t(\mathbf{H}_t)) + \lambda_{N} || f_t(\mathbf{H}_t)||^2
    \Bigg],
\end{align*}
where $(\hat{d}^*_{t+1}, \dots \hat{d}^*_T)$ are obtained prior to stage $t$, and the $T$-stage estimator does not account for treatments followed afterwards, i.e., $\prod_{\tau = T+1}^T\mathbb{I}[A_\tau = \hat{d}^*_\tau(\mathbf{H}_\tau)] \doteq 1$.

The second approach allows simultaneous estimation for all stages. In a two-stage problem, for example, the SOWL estimator, say $\hat{f}^*_S$, is defined as follows:
\begin{align*}
    \hat{f}^*_S \doteq \argmax_{f \in \mathcal{F}} \mathbb{P}_N \Bigg[
    &\frac{Y \psi(A_0 f_0(H_0), A_1 f_1(\mathbf{H_1})) }{\prod_{t=0}^1\pi_t(A_t|\mathbf{H_t})} - \lambda_N (||f_0(H_0)||^2 + || f_1(\mathbf{H_1})||^2)
    \Bigg],
\end{align*}
with $\psi(x_1, x_2) \doteq \min(x_1-1, x_2-1,0)+1$ being a concave surrogate for the product of the two (discontinuous) indication functions, introduced to limit computational issues.

Even if numerical examples show that BOWL and SOWL have superior performances compared to existing direct methods, significant information is lost as $t$ decreases, since only subjects who followed the estimated optimal regime after stage $t$ are used in the backward optimization algorithm. To overcome this problem, an augumented BOWL estimator, leveraging the use of Q-functions, has been proposed and we refer to the original work of \cite{liu2018augmented} for this extension.

\begin{table}[ht]
\centering
    \begin{tabularx}{\textwidth}{r|X}
        Reference \& Acronym & Extension type\\
        \hline \hline
        \cite{zhao2015new}\\ BOWL + SOWL & \textbf{Extension to $T$-stages}, with $T<\infty$. The authors proposed two methods: one performs an iterative backward OWL (BOWL) estimation, the other a simultaneous OWL (SOWL) estimation.\\
        \hline
        \cite{liu2018augmented}\\ AOL & \textbf{Extension to negative outcomes and multiple stages}. The authors proposed an augumented version for the weight of the OWL (AOL) integrating OWL and Q-functions. The robust augmentation, making use of predicted pseudo-outcomes from regression models for Q-functions, reduces the variability of weights and improves estimation accuracy.\\
        \hline
        \cite{zhou2017residual}\\ RWL & \textbf{Extension to continuous, binary, and count outcomes, and possibility of variable selection}. The authors proposed a general framework, called Residual Weighted Learning (RWL), which employs a \textit{smoothed ramp loss} and derived outcome residuals with a regression model.\\
        \hline
        \cite{chen_estimating_2018}\\ GOWL & \textbf{Extension to ordinal treatments and negative outcomes}. The authors proposed a generalized OWL (GOWL) based on a modified loss function and a reformulation of the objective function in the standard OWL.\\
        \hline
        \cite{zhang_multicategory_2020}\\ MOML & \textbf{Extension to multicategory treatment scenarios and negative outcomes}. The authors used sequential binary methods employing margin-based learning (based on a \textit{large-margin unified machine loss}), which has a special case the standard OWL.\\
        \hline
        \cite{fu_robust_2019}\\ ROWL & \textbf{Extension to outliers, multicategory treatments, and negative outcomes}. The authors proposed a robust OWL (ROWL), based on an angle-based classification structure, designed for multicategory classification problems, and a new family of \textit{robust loss} functions to build more stable DTRs.\\
        \hline
    \end{tabularx}
    \caption{Extensions of the standard OWL estimator for DTRs.}
    \label{tab: OWL_extensions}
\end{table}

\bigskip
More recently, under both the direct weighted classification and the indirect blip function frameworks, \cite{luedtke_super-learning_2016} and colleagues~\citep{montoya_optimal_2023} introduced the \textit{SuperLearner} ensemble method~\citep{van_der_laan_super_2007} in the DTR arena. Rather than a-priori selecting an estimation framework and algorithm, estimators from both frameworks (and a user-supplied library of candidate algorithms), are combined by using a super-learning based cross-validation selector that seeks to minimize an appropriate cross-validated risk. The full approach is described in Section 3.3 of~\cite{montoya_optimal_2023}.
\bigskip

\noindent\textbf{V-learning for indefinite-horizon problems} Most of the work in DTRs focuses on the finite-horizon setting with a very limited literature addressing the in(de)finite case. Yet, for some chronic conditions or those with short or nonfixed time intervals, it may be more natural to assume an in(de)finite time horizon. Tackling this specific setting, \cite{ertefaie2018constructing} proposed an indirect Q-learning approach, while \cite{luckett2020estimating} focused on searching an optimal policy over a prespecified class of policies, as in direct methods. Motivated by an mHealth application, where policy estimation is continuously updated in real time as data accumulate (and starting with small sample sizes), \cite{luckett2020estimating} introduced \textit{V-learning}. The objective is represented by the value function, written as:
\begin{align} \label{eq: Luckett_value}
    V^d(x_t) = \sum_{\tau \geq t} \mathbb{E} \left[ \gamma^{\tau-t}Y_{\tau+1} \left( \prod_{v = t}^\tau \frac{d(A_v|X_v)}{\pi_v(A_v| S_v)}\right) \middle | X_t = x_t \right],
\end{align}
with $\pi$ an exploration policy, which can be seen as the randomization probability in a randomized trial, and $d$ an arbitrary policy which we want to learn about. Under a time-homogeneous Markov assumption, and provided interchange of the sum and integration is justified, 
for any function $\psi$ defined on the state space $\mathcal{X}_t$, the value function in Eq.~\eqref{eq: Luckett_value} satisfies an importance-weighted variant of the Bellman optimality given by: 
\begin{align*}
    0 = \mathbb{E} \left[\frac{d(A_t|X_t)}{\pi_t(A_t| S_t)} \left(Y_{t+1} + \gamma V^d(X_{t+1}) - V^d(X_t)\right)\psi(X_t) \right].
\end{align*}

Let now $V^d(x;\theta)$, with $ \theta \in \Theta \subseteq \mathbb{R}^q $, be a model for $V^d(x)$. Assuming that $V^d(x;\theta)$ is differentiable everywhere in $\theta$, for fixed $x$ and $d$, and denoted with $\psi(x) \doteq \nabla_{\theta} V^d(x; \theta)$, the proposed estimating equation function is given by:
\begin{align*}
    \hat{\Lambda}(\theta) = \mathbb{P}_N \Bigg[\sum_{t=0}^T \frac{d(A_t|X_t)}{\pi_t(A_t| S_t)} \Bigg( Y_{t+1} + \gamma V^d(X_{t+1}; \theta)- V^d(X_t; \theta) \Bigg) \nabla_{\theta} V^d(X_t; \theta)\Bigg].
\end{align*}

An estimate $\hat{\theta}$ can be obtained by minimizing $\hat{M}(\theta) \doteq \hat{\Lambda}(\theta)^T\hat{W}^{-1}\hat{\Lambda}(\theta) + \lambda \mathcal{P}(\theta)$, with $\hat{W}$ a positive definite matrix in $\mathbb{R}^{q\times q}$, $\lambda$ a tuning parameter and $\mathcal{P}: \mathbb{R}^q \to \mathbb{R}_+$ a penalty function. 
The optimal estimate $\hat{d}^*$ is then the argmax of $V^d(x;\hat{\theta})$.

\subsection{Data sources for constructing DTRs} \label{Sec: Data_sources}

For the study of DTRs, three main sources of data are typically considered in the literature: i) longitudinal observational studies including cohort studies and EHRs, ii) sequentially randomized studies, with \textit{sequential multiple assignment randomized trial} (SMART) designs~\citep{lavori_dynamic_2004,murphy_experimental_2005} being the ``gold standard'', and iii) dynamical system models, a tool transposed from control engineering~\citep{rivera_using_2007}. While most real-life studies are based on the first type of data, experimental data sources represent the highest-quality data source for developing DTRs. The third approach has been more peripheral within the DTR literature, typically confined to methodological works aiming at developing and improving existing methodologies. Despite their artificial nature, dynamical systems are built according to biological, behavioral, or social models that simulate realistic individual trajectories. Clinical examples are provided in~\cite{thall2007bayesian,rosenberg_using_2007}, and more behavioral-oriented cases can be found in \cite{rivera_using_2007,navarro-barrientos_dynamical_2011}. For other illustrative examples, we refer to \cite{deliu_dynamic_2022}. In what follows, the first two types of data sources are discussed.
\bigskip

\noindent \textbf{Longitudinal observational studies} Observational data for the study of DTRs include longitudinal trajectories arising from EHRs and other administrative databases or cohort studies~\citep{rosthoj_estimation_2006,van_der_laan_statistical_2007}. They represent a major data source in the biomedical field, constituting an appealing option in scenarios in which a trial would be either cost prohibitive or of concern from an ethical or logistic perspective (e.g., in several chronic diseases such as diabetes or HIV). For this reason, as discussed in~\cite{mahar_scoping_2021}, the estimation of optimal DTRs using observational data has been most concentrated in the area of HIV/AIDS (27, 43\%), followed by cancer (8, 13\%), and diabetes (6, 10\%). 

Compared to experimental data arising from randomized clinical trials, using observational data to construct DTRs provides the advantage of evaluating a wider range of treatments at a lower cost. Furthermore, they allow data collection over continuous and indefinite time horizons. However, since the treatments are not randomized within the study, the procedures for drawing causal inference may be affected by the potential presence of time-varying confounders. In particular, the reasons why different individuals receive different treatments or the reasons why one individual receives different treatments at different times are not known with certainty.
\bigskip

\noindent \textbf{Sequentially-randomized studies} Although observational data offer a cost-acceptable option and reflect the population's heterogeneity, they present several challenges that make estimation challenging and often subject to various hidden biases. Therefore, randomized data, when available, are preferable for more accurate estimation and stronger statistical inference~\citep{rubin1974estimating, rosenberger2019randomization}. This is especially important when dealing with DTRs since hidden biases can compound over stages. 
Randomized trials, and most interestingly, SMART designs, are the ``gold standard'' in DTRs. Randomization coupled with compliance allows causal interpretations to be drawn from statistical association, and are the most effective designs in these multistage medical settings. 
    
\begin{figure}[htb]
    \centering
    \includegraphics[width=.9\linewidth]{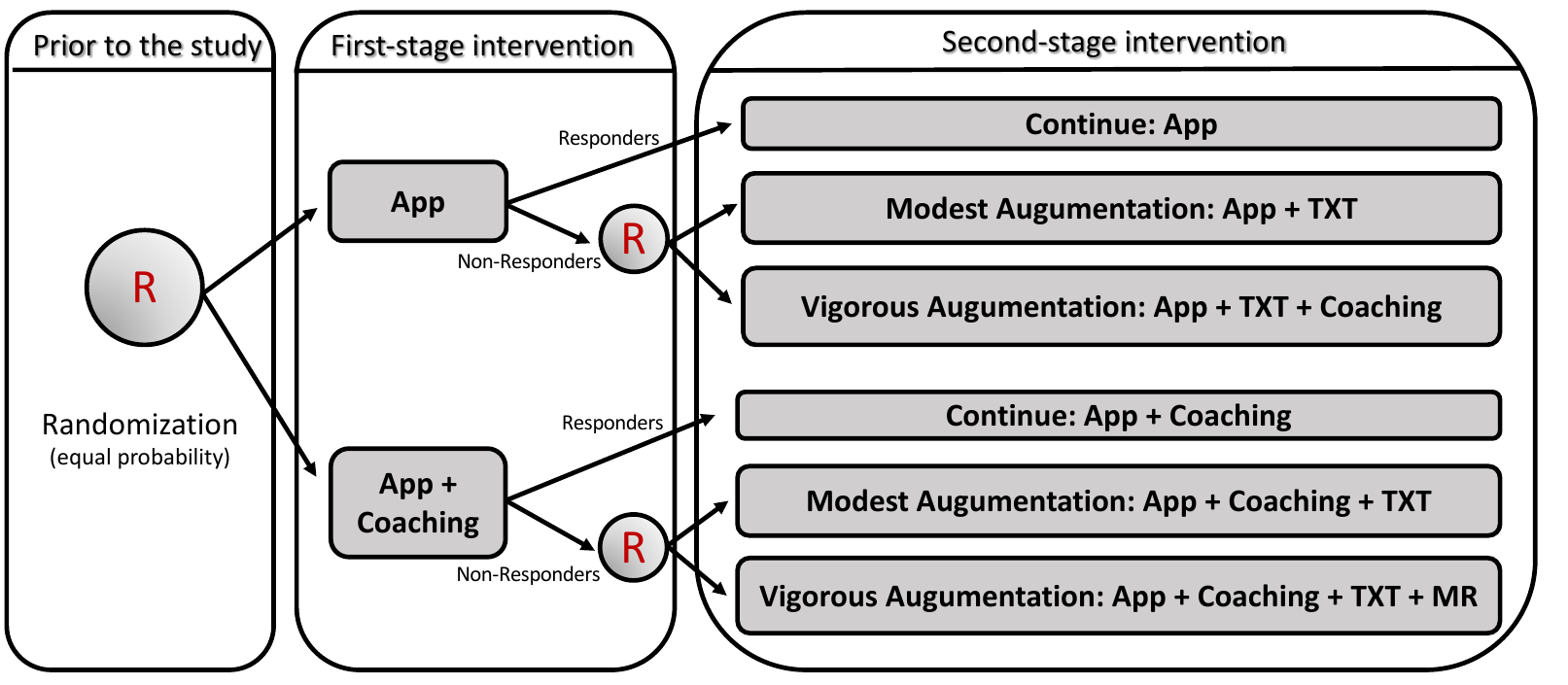}
    \caption{Schematic of the SMART design of the weight loss management study in~\cite{pfammatter_smart_2019}. App denotes a mobile app, TXT a support text message, and MR meal replacement. The response is defined as a weight loss of at least $0.5$ lb on average per week.}
    \label{fig: smart-wl}
\end{figure}

A SMART design is characterized by multiple stages of treatment, each stage corresponding to one of the critical decision time point in which randomization occurs. At each subsequent stage, rerandomizations may depend on information collected after previous treatments, but prior to assigning the new treatment, e.g., how well the patient responded to the previous treatment. On the basis of the extent of multiple randomizations, different types of SMARTs can be defined. These include SMARTs in which only nonresponders are rerandomized, and SMARTs in which both responders and nonresponders are rerandomized. In addition, randomization could be made only for one of the initial treatments or for all initial treatments.
For a concrete example, see Figure~\ref{fig: smart-wl} for the schematic of the SMART design adopted for the weight loss management study in~\cite{pfammatter_smart_2019}. This SMART example involves two stages of treatment and/or experimentation; in general, it may involve as many stages as practically feasible. 

\subsection{Case study: PROJECT QUIT – FOREVER FREE} \label{Sec: DTR_applications_challenges}

Based on a two-stage SMART design, the \textit{PROJECT QUIT – FOREVER FREE} study aimed to develop/compare internet-based (precursors to mobile-based) behavioral interventions for smoking cessation and relapse prevention. The primary aim, based on the six-month-long first stage of the study, i.e., the \textit{PROJECT QUIT}, was to find an optimal multi-factor behavioral intervention to help adult smokers quit smoking; see~\cite{strecher_web-based_2008} for more details. The second stage, known as \textit{FOREVER FREE}, was a six-month-long follow-on study to help \textit{PROJECT QUIT} participants who quit remain non-smoking, and offer a second chance to those who failed to give up smoking at the previous stage. These two stages were then considered together with the goal of finding an optimal DTR over a twelve-month study period; this was a secondary aim of the main study. RL was not used in the design phase; in other words, this was not an instance of \textit{online learning}. The RL-type learning happened \textit{offline} on completion of data collection, when Q-learning with linear model and a variant (\textit{soft-thresholding}) were employed. The choice of Q-learning was driven by its simplicity and interpretability. Detailed results from this secondary analysis can be found in~\cite{chakraborty_study_2009} and \cite{chakraborty2010inference}. Here, we only summarize the characteristics of the study in relation to the RL framework and the main challenges we faced.

\begin{description}
\item[\textbf{Selection of interventions and tailoring variables}] In \textit{PROJECT QUIT}, the original plan was to randomly administer and test six behavioral intervention components (factors), each varied at two levels (highly individually tailored vs. not), according to a 32-cell \textit{fractional factorial design} (FFD). However, due to a program error, one of those factors was not properly implemented. Subsequently, utilizing the factorial structure, the design was ``folded" to convert it to a 16-cell FFD and that particular factor was removed from further consideration at the analysis stage. In the primary analysis~\citep{strecher_web-based_2008}, which was a traditional logistic regression analysis, only two of the five factors were statistically significant. Based on this finding, only these two intervention factors (each at two levels) were considered in the stage-1 Q-learning model. Likewise, various participant-level contextual/tailoring variables were considered in the primary analysis, but only three of them (education, motivation, and self-efficacy) were statistically significant. Again, informed by the primary analysis, only these variables were considered in the stage-1 model of Q-learning, allowing a parsimonious choice of model. In \textit{FOREVER FREE}, originally there were four versions of an active behavioral intervention and a control arm, i.e., five arms in total. But later in the analysis stage, the four versions of the active intervention were found to be minimally different from each other, and hence collapsed. This decision resulted in only two intervention arms at the second stage of Q-learning. 
\item[\textbf{Reward function}] The primary outcomes at both stages of the original study were the corresponding seven-day point prevalence of smoking (i.e., whether or not the participant smoked even a single cigarette in the last seven days at six months following the randomization), a dominant measure in the smoking cessation literature. These outcomes were considered as the stage-specific reward functions in Q-learning~\citep{chakraborty2010inference}. However, the basic operationalization of Q-learning is for continuous outcomes, while the seven-day point prevalence outcomes were binary. Additional Q-learning analysis with a relatively more continuous reward function, the number of months not smoked in the last six months (a secondary outcome in the main study), was also conducted~\citep{chakraborty_study_2009}. Qualitatively, the results were not too different.
\item[\textbf{Missing data in the reward variable}]
In \textit{PROJECT QUIT}, 1848 participants were randomly allocated to various interventions, but only 479 of them decided to continue to \textit{FOREVER FREE}; this flexibility was part of the protocol, and hence the remaining \textit{PROJECT QUIT} participants were not considered to be drop-outs for the \textit{FOREVER FREE} part of the study. However, only 1401 out of 1848 stage-1 participants completed the six-month outcome survey; these
1401 participants were treated as complete cases, while the remaining 447 participants were considered drop-outs in stage 1. Similarly, 281 participants (out of 479) who completed the stage-2 six-month survey were treated as complete cases, while the remaining 198 participants were considered drop-outs in stage 2. Descriptive checks revealed that drop-out was more or less uniform across the different intervention arms at both stages. One can employ modern missing data analysis techniques, e.g., \textit{multiple imputation}, before applying Q-learning or other offline RL methods on SMART data to learn about optimal DTRs~\citep[see][for details]{shortreed_multiple_2014}. In the case of the \textit{PROJECT QUIT - FOREVER FREE} data, \cite{chakraborty2010inference} only presented a complete-case analysis, while \cite{chakraborty_study_2009} also presented Q-learning analysis of multiply-imputed data.   
\item[\textbf{Statistical inference}] In DTRs, inference is complicated by the presence of nonregularity~\citep{robins2004optimal}, a phenomenon characterizing the lack of locally uniform convergence. This can be a result of the sampling distributions of the corresponding estimators changing abruptly as a function of the true underlying parameters. In DTRs, this may occur when two or more treatments produce (nearly) the same mean optimal outcome. 
To solve this problem, \cite{chakraborty2010inference} proposed, first, two alternative ways of shrinking or thresholding values near zero, and then a general method for bootstrapping under nonregularity, i.e., \textit{m-out-of-n bootstrap}~\citep{chakraborty2013inference}. We refer to these references for the readers interested in this problem.

\end{description}

\section{Just-in-time Adaptive Interventions in Digital Health} \label{Sec: JITAIs}

Digital and mobile health technologies hold the great potential to deliver just-in-time adaptive interventions (JITAIs), i.e., personalized interventions continuously adapted to the real-time contexts of users. As in DTRs, JITAIs represent a sequence of decision rules tailored to individual users. Their peculiarity is in providing interventions according to the user's in-the-moment context or needs, e.g., time, location, or current activity, including considerations about whether and when the intervention is needed. This feature is enabled by mobile technologies (e.g., wearable devices, accelerometers, or smartphones) that collect data on a continuous basis allowing one to adapt intervention components in real time. In such settings, learning typically occurs online over in(de)finite horizons, and the target policy corresponds to the exploration or behavioral policy. Unlike DTRs, the number of decision points in JITAIs can be hundreds or even thousands, and the intervention can be delivered each minute, hour, or day. This distinctive feature, at least partially, contributed to their increasing popularity in a variety of behavioral domains, including physical activity~\citep{hardeman2019systematic, figueroa_daily_2022}, addictive disorders in alcohol and drug use \citep{goldstein2017return, garnett2019development, bell_notifications_2020}, smoking cessation \citep{naughton2017delivering} and obesity/weight management \citep{aswani_behavioral_2019}.
Furthermore, JITAIs have the potential to improve access to quality care in underserved communities and thus alleviate health disparities, a significant public health concern~\citep{liu_microrandomized_2023}.

Given the high granularity of the JITAI data, two outcome components play a determinant role in the adaptation and optimization of interventions in real time~\citep{nahum-shani_introduction_2019}:
\begin{enumerate}
    \item[(v)] the \textbf{proximal outcome(s)} $\{Y_t\}_{t>0}$ directly targeted by an intervention, easily observable and expected to influence a longer-term outcome of interest, according to some mediation theory~\citep{mackinnon_mediation_2007};
    \item[(vi)] the \textbf{distal outcome(s)}, representing the long-term health outcome of interest (typically clinical) and ultimate goal of the overall AI.
\end{enumerate}
Taken together with the four components that define a DTR (see Section~\ref{Sec: DTRs}), they represent the six key ingredients of a JITAI. Note that the proximal outcomes can also be used as tailoring variables for guiding later-stage decisions. Unlike DTRs--which target the distal outcome and may or may not have an intermediate (proximal) outcome--in JITAIs, the proximal outcomes represent the direct and in-the-moment target of the intervention. The distal outcome is expected to improve only on the basis of domain knowledge about its relationship with proximal outcomes, but it is not formally included in an optimization problem. A more detailed discussion of the differences between DTRs and JITAIs is provided in~\cite{deliu2023reinforcement}.

\subsection{RL methods for JITAIs in digital health}
JITAIs are carried out in dynamic environments where the context and needs of individual users can change rapidly~\citep{nahum2015building, nahum2018just}. Therefore, methodologies for delivering JITAIs are required to perform almost continuous learning--with no definite time horizon--and to provide interventions \textit{online} as data accumulate, often utilizing trajectories defined over very short time periods. Note that in such settings, the exploration policy $\boldsymbol{\pi}$ used to collect the samples corresponds to the target policy $\boldsymbol{d}$ we want to improve and optimize; that is, $\boldsymbol{\pi} = \boldsymbol{d}$. However, existing methods for DTRs mainly target a finite-time horizon problem and are implemented offline with backward induction (as in Q-learning); therefore, they are not directly applicable. Furthermore, by carrying over an entire history of an individual, they may not be feasible from a computational perspective. In such problems, a simplified RL framework, known as the multi-armed bandit (MAB) problem, represents an attractive approach and is increasigly being used within the digital and mobile health domains~\citep{tewari2017ads}. In what follows, after an illustration of the MAB framework, we describe some popular MAB algorithms that have been used in digital health.

\subsubsection{The multi-armed bandit framework} \label{Sec: MABs}
MAB problems can be viewed as a subclass of RL problems \citep{sutton18_rl}. In the simplest stateless case, the environment does not have any state transitions and actions can be determined according to a single-stage decision-making framework. 
Generally speaking, the MAB problem (also called the $K$-armed bandit problem) is a problem in which a limited set of resources (e.g., a group of individuals) must be allocated between competing choices in order to maximize the total expected reward over time. Each of the $K$ choices (i.e., \textit{arms} or actions) provides a different reward, whose probability distribution is specific to that choice. If one knew the expected reward (or value) of each action, then it would be trivial to solve the bandit problem: they would always select the action with the highest value. However, as this information is only partially gained for the selected actions, at each decision time $t$ the agent must trade-off between optimizing its decisions based on acquired knowledge up to time $t$ (\textit{exploitation}) and acquiring new knowledge about the expected rewards of the other actions (\textit{exploration}). The problem conforms to the class of on-policy RL, where the same policy $\boldsymbol{\pi}$ used to explore the actions is evaluated and improved throughout the learning process. For this reason, within this section, we avoid a distinction between $\boldsymbol{\pi}$ and $\boldsymbol{d}$, and we use $\boldsymbol{\pi}$ to refer to both the target policy and the exploration policy. 

MAB problems can incorporate some context, which in mapped into appropriate interventions or arms (\textit{contextual} MABs), or solve a context-free task (\textit{non-contextual} MABs), where no side-information is used. In the theory of sequential decision making, contextual MABs occupy a middle ground between non-contextual MABs~\citep{bubeck2012regret, auer2002nonstochastic} and full-blown RL. 

The most typical assumption is that contexts $\{X_t\}_{t \in \mathbb{N}}$ are independent and identically distributed (i.i.d.) with some fixed but unknown distribution. This means that action $A_t$ at time $t$ has an \textit{in-the-moment} effect on the proximal reward $Y_{t+1}$ at time $t+1$, but not on the distribution of future rewards $\{Y_\tau\}_{\tau \geq t+2}$, for which the i.i.d. property also holds. Under this assumption, one can be completely \textit{myopic} (with $\gamma = 0$ in Eqs.~\eqref{eq: return} and~\eqref{eq: bellman_opt_Q}) and ignore the effect of an action on the distant future in searching for a good policy. 
In such contextual MABs, and further in the context-free MAB problem, the trajectory distributions are simplified as follows:
\begin{align*}
P_{\boldsymbol{\pi}} &\doteq 
p_0(x_0) \prod_{t \geq 0} \pi_{t}(a_{t} |x_t)p_{t+1}(x_{t+1}, y_{t+1}|x_t, a_{t})\quad \quad \quad \quad\quad\quad \ \ \text{[Contextual MAB]} \\ &= p_0(x_0) \prod_{t \geq 0} \pi_{t}(a_{t} |x_t)
p_{t+1}(x_{t+1}) r_{t+1}(y_{t+1}|x_t, x_{t+1}, a_{t})\\
P_{\boldsymbol{\pi}} &\doteq 
\prod_{t \geq 0} \pi_{t}(a_{t})r_{t+1}(y_{t+1}|a_{t}).\quad\quad\quad\quad\quad\quad\quad\quad\quad\quad\quad\quad \text{[Non-contextual MAB]}
\end{align*}

As in the general RL problem, the goal of an MAB problem is to select the optimal arm at each time $t$ to maximize the expected return, alternatively (and with a slightly different nuance) expressed in the bandit literature in terms of minimizing the \textit{total regret}. 
Formally, denoted by
$ A_t^* \doteq \argmax_{a_t \in \mathcal{A}}\mathbb{E}(Y_{t+1} | X_t = x_t, A_t = a_t)
$
the optimal arm at time $t$, we define the \textit{immediate regret} $\Delta(A_t)$ of action $A_t$ as the difference between the expected reward of the optimal arm $A_t^*$ and the expected reward of the ultimately chosen arm $A_t$, i.e.,
\begin{align} \label{eq: regret}
    \Delta(A_t) \doteq \mathbb{E}(Y_{t+1} | X_t, A_t^*) - \mathbb{E}(Y_{t+1} | X_t, A_t).
\end{align}

A nonexhaustive correspondence table between the MAB and JITAI terminologies is reported in \textbf{Table~\ref{tab: terminology2}}.
\begin{table}[htb]
\centering
    \caption{Terminology correspondence between MABs and JITAIs.}
    \begin{tabularx}{\textwidth}{l|X|X}
    \hline
        \multicolumn{1}{c|}{\textbf{Notation}} &\multicolumn{2}{c}{\textbf{Terminology}}\\
        & MABs & JITAIs in mHealth \\
        \hline 
        $i$& Trajectory & User, Subject, Individual\\
        $t$& Round, Step & Time, Round, Step\\
        $X$& Context& Context, Contextual Variables\\
        $A$& Arm & Intervention, Arm\\
        $Y$& Reward, Payoff & Proximal Outcome\\
        $\mathbf{H}$ & Filtration & Filtration\\
        $\boldsymbol{\pi}$/$\mathbf{d}$& Policy & Just-in-time Adaptive Intervention\\
        \hline
    \end{tabularx}
    \label{tab: terminology2}
\end{table}

With a few exceptions, the contextual MAB algorithms applied in mHealth are based on and rely on the field-specific adaptation of two fundamental contextual bandit approaches: the Upper Confidence Bound \citep[UCB;][]{li2010contextual, chu2011contextual} and the Thompson sampling~\citep[TS;][]{thompson_likelihood_1933, agrawal2013thompson} strategies. Exceptions include the Actor-Critic strategy used e.g., in~\cite{greenewald17}. 
\bigskip

\noindent\textbf{Contextual bandits with linear UCB} Linear Upper Confidence Bound (LinUCB) bandits~\citep{li2010contextual, chu2011contextual} represent an extension of the UCB algorithm \citep{auer2002finite} for MAB problems to contextual MAB problems. 
It assumes that the expected reward is a linear function of the context-action feature $f(X_t, A_t) \in \mathbb{R}^{d'}$, i.e., $\mathbb{E}[Y_{t+1} | X_t, A_t] = f(X_t, A_t)^T\mu$, with $\mu \in \mathbb{R}^{d'}$ an unknown reward parameter. In this work, we refer to general features~\citep[constructed e.g., via linear basis, polynomials or splines expansion; see e.g.,][]{marsh2001spline} rather than a standard linear function that may not capture nonlinearities in the data. Similar modeling considerations have been made in Q-learning (by using, e.g., DNNs).

At each time $t$, revealed the context $X_t$, LinUCB calculates the upper confidence bound for the expected reward for all possible actions and then selects the action associated with the highest UCB. Denoted by $U_t(a_t)$ the UCB of arm $a_t$ at time $t$, \cite{li2010contextual} and \cite{chu2011contextual} proposed the formulation:
\begin{align*} 
    \hat{U}_t(a_t) \doteq \mathbb{E}[Y_{t+1} | X_t, A_t] + \alpha s_t(a_t) = f(X_t = x_t, A_t = a_t)^T \hat{\mu}_{t} + \alpha s_t(a_t),
\end{align*}
where $\hat{\mu}_{t}$ is an estimator of the unknown regression coefficient $\mu_t$ and $s_t(a_t)$ is defined as $\sqrt{f(X_{t}, A_t = a_t)^T B_{t}^{-1}f(X_{t}, A_t = a_t)}$, with $B_t \doteq \lambda \mathbb{I}_{d'} + \sum_{\tau=0}^{t-1}f(X_\tau, A_\tau = \tilde{a}_\tau)f(X_\tau, A_\tau = \tilde{a}_\tau)^T$. $B_t$ is computed recursively at each time step $t$, by taking into account the context-action features associated with the optimal actions $\left\{\tilde{a}_\tau \doteq \argmax_{a_\tau \in \mathcal{A}}U_\tau(a_\tau)\right\}_{\tau = 0,1,\dots,t-1}$ estimated at previous rounds.
Note that the first part $f(X_t, A_t = a_t)^T \hat{\mu}_{t}$ reflects the current estimate of the reward, while the second part $s_t(a_t)$ is an indication of its uncertainty; thus, it naturally balances between exploration and exploitation.
The tuning parameter $\alpha > 0$ balances the trade-off between exploration and exploitation: small values of $\alpha$ favor exploitation, while larger values of $\alpha$ favor exploration.

Moving from pure bandit and statistical theory to real-world digital health applications, the use of LinUCB has been reported in \cite{forman2019can}, among others. Here, in the context of behavioral weight loss and maintenance, a pilot study has been conducted to evaluate the feasibility and acceptability of an RL-based intervention. Participants were randomized into a nonoptimized, an individually optimized (individual reward maximization), and a group optimized (group reward maximization) group. The study showed the advantages of the RL-based optimized groups in terms of the outcome of interest, not only being feasible to deploy and acceptable to participants and coaches, but also achieving desirable results at roughly one-third the cost.
\bigskip

\noindent\textbf{Contextual bandits with linear Thompson sampling} Under the same linear assumption of LinUCB, \cite{agrawal2013thompson} proposed a randomized version of the latter, based on a generalization of the TS technique for i.i.d. contextual MAB problems. Based on the Bayesian framework, the idea of TS is to randomly allocate each arm according to its posterior probability of being optimal. More specifically, assuming a Gaussian prior for the $\mu$ parameter $\mu \sim \mathcal{N}(\mathbf{0}_{d'}, \nu^2 \mathbb{I}_{d'})$ and a Gaussian distribution for the reward $Y_t|\mu,f(X_t, A_t) \sim \mathcal{N}(f(X_t, A_t)^T\mu, \nu^2)$, for some $\nu > 0$, at each time $t$ the optimal arm $\tilde{a}_t$ is the one that maximizes the `a-posteriori' estimated expected reward, i.e., $f(X_t, A_t)^T\tilde{\mu}_t$. The posterior nature is reflected in $\tilde{\mu}_t$, which represents a sample from the posterior distribution, computed recursively and given by $\mathcal{N}(\hat{\mu}_t, \nu^2 B_t^{-1})$, with $\hat{\mu}_t \doteq B_t^{-1}b_t$, where $B_t \doteq \mathbb{I}_{d'} + \sum_{\tau=0}^{t-1}f(X_\tau, A_\tau = \tilde{a}_\tau)f(X_\tau, A_\tau = \tilde{a}_\tau)^T$ and $b_t \doteq \sum_{\tau=0}^{t-1}f(X_\tau, A_\tau = \tilde{a}_\tau)Y_{\tau+1}(X_\tau, A_\tau = \tilde{a}_\tau)$. The policy $\pi$ at each time $t$ is thus explicitly defined as:
\begin{align*}
     \pi_t(a) &= 
     \mathbb{P}\Big(\mathbb{E}[Y_{t+1} \mid X_t = x_t, A_{t} = a] \geq \mathbb{E}[Y_{t+1} \mid X_t = x_t, A_{t} = a'], \forall a' \neq a  \mid \mathcal{F}_{t-1} \Big),  
\end{align*} 
where the conditioning term $\mathcal{F}_{t-1}$ reflects the posterior nature of this strategy. 

Given all the trajectory information up to time $t$, $\mathcal{T}_{t-1} = \{(X_\tau, A_\tau, Y_{\tau+1})\}_{\tau = 0,1,\dots, t-1}$ and $f(X_t, A_t)$, LinUCB is deterministic and allows exploration through the uncertainty term $\alpha s_t(a_t)$. On the other hand, in TS, exploration is given by the random draws from the posterior distribution. Note that the standard deviation $\alpha s_t(a_t)$ characterizing LinUCB has the same order as the standard deviation of the posterior distribution of the reward $Y_t|\mu_t,f(X_t, A_t) \sim \mathcal{N}(f(X_t, A_t = a_t)^T\hat{\mu}_t, \nu^2 f(X_t, A_t = a_t)^TB_t^{-1}f(X_t, A_t = a_t))$ used in TS, where $ f(X_t, A_t = a_t)^TB_t^{-1}f(X_t, A_t = a_t) = s_t(a_t)$ by definition.

\bigskip

\noindent\textbf{Actor-critic contextual bandits} Specifically addressing personalized mHealth intervention problems, \cite{lei2016online} proposed to use a particular RL setting, called actor-critic~\citep{grondman_survey_2012}, based on which both policies and value functions are learned. The ``actor'' is the component that learns policies, and the ``critic'' is the component that learns about whatever policy the actor is currently following to ``criticize'' its choices \citep{sutton18_rl}.

Assuming a linear model for the reward $Y_{t+1} = f(X_t, A_t)^T\mu_t + \epsilon_{t+1}$, with $f(X_t, A_t) \in \mathbb{R}^{d'}$, $\mu \in \mathbb{R}^{d'}$ and i.i.d. error terms $\{\epsilon_t\}_{t \in \mathbb{N}}$ with mean $0$ and variance $\sigma^2$, and taking into account a binary action space $\mathcal{A} = \{ 0, 1\}$, \cite{lei2016online} formulated an online policy learning procedure as a contextual bandit problem, and proposed a class of parametrized stochastic policies with $\mathbb{P}(A = 1|X = x) = \pi(1|x; \theta) = \frac{e^{g(x)^T\theta}}{1+e^{g(x)^T\theta}}$, and $g(x)$ a $p$-dimensional policy feature.

To maintain variety of treatment and increasing engagement, a stochastic chance constraint related to a parametrized stationary policy is introduced. In this specific case, if we denote with $\{\pi(a|x; \theta): \theta \in \Theta \subseteq \mathbb{R}^d\}$ the class of parameterized stochastic policies, the stochasticity constraint has the form:
\begin{align}\label{constraint}
    \mathbb{P}(\pi_{min} \leq \pi(A = 1|X; \theta) \leq 1-\pi_{min}) \geq 1- \alpha,
\end{align}
with $\pi_{min} \in (0, .5)$ and $\alpha \in (0,1)$ controlling the amount of stochasticity.

An optimal policy can then be obtained by maximizing the expected reward under the policy $\pi(a|x; \theta)$, i.e., $V^\pi(\theta) \doteq \mathbb{E}_{\pi_\theta}(Y)$, subject to the constraint in Eq.~\eqref{constraint}. 
Solving this constrained optimization problem involves a major difficulty given the nonconvex constraint on $\theta$, which involves also some nonsmoothness. To circumvent this difficulty, the authors relax the above constraint by bounding the probability in Eq.~\eqref{constraint} using Markov's inequality, and then solving the constrained optimization problem using the Lagrangian function $J_\lambda(\theta)$, with $\lambda$ the Lagrangian multiplier. That is, for a fixed $\lambda$, the optimal policy $\pi^* \doteq \pi_{\theta^*}$ is the one with $\theta^*$ given by:
\begin{align*}
    \theta^* \doteq \argmax_{\theta \in \Theta} J_\lambda(\theta),
\end{align*}
where $J_\lambda(\theta)$, also referred to as regularized average reward, is defined as
\begin{align*}
    J_\lambda(\theta) &\doteq V^\pi(\theta) - \lambda \theta^T \mathbb{E}(g(X)g(X)^T)\theta\\
    & = \mathbb{E}_{\pi_\theta}(Y) - \lambda \theta^T \mathbb{E}(g(X)g(X)^T)\theta\\
    & = \mathbb{E}_{p(x)}\mathbb{E}_{\pi(a|x; \theta)}[E(Y|X = x; A = a)] - \lambda \theta^T \mathbb{E}(g(X)g(X)^T)\theta.
\end{align*}

Being $J_\lambda(\theta)$ unknown, its MC version is considered:
\begin{align*}
    \widehat{J}_\lambda(\theta) = \mathbb{P}_N \left[ \sum_{a}E(Y|X = x; A = a)\pi(a|X = x; \theta) - \lambda \theta^T (g(x)g(x)^T)\theta \right],
\end{align*}
where $\mathbb{P}_N$ denotes the empirical average on $N$ i.i.d. samples.

For estimating the expected reward $E(Y|X = x; A = a)$, the linear assumption $\mathbb{E}(Y_{t+1}|X_t = x, A_t = a) = f(x, a)^T\mu$ is used, and the $L_2$ norm penalized least square estimator is considered. This helps to overcome the full rank requirement of the matrix $\sum_{t = 0}^T f(X_t, A_t)f(X_t, A_t)^T$ at the beginning of the experiment when running the algorithm online.

While the proposed algorithm is formulated for a binary action space, we notice that this class of methods has been originally introduced to overcome the limitation of methods that fail to address complex action space problems (e.g., tabular Q-learning). Thus, the greater advantage of an actor-critic approach occurs in settings with action spaces more complex than the binary case, which may be solved by simpler methods such as LinTS or LinUCB.

\subsection{Data sources for building JITAIs in digital health}
Typical experimental designs for building JITAIs are represented by \textit{factorial experiments}~\citep{collins_comparison_2009}, or, most notably, \textit{micro-randomized trials}~\citep[MRTs;][]{klasnja2015microrandomized}. In MRTs, individuals are randomized hundreds or thousands of times over the course of the study, and, in a typical multicomponent intervention study, the multiple components can be randomized concurrently, making micro-randomization a form of a sequential factorial design. The goal of these trials is to optimize mHealth interventions while assessing the causal effects of each randomized intervention component and evaluating whether the intervention effects vary with time or the current context of individuals. A review of this cutting-edge design, covering both its classical variants and its adaptive counterpart, together with the associated statistical challenges, is presented in~\cite{liu_microrandomized_2023}.

\begin{figure*}
    \centering
    \includegraphics[width=1\linewidth]{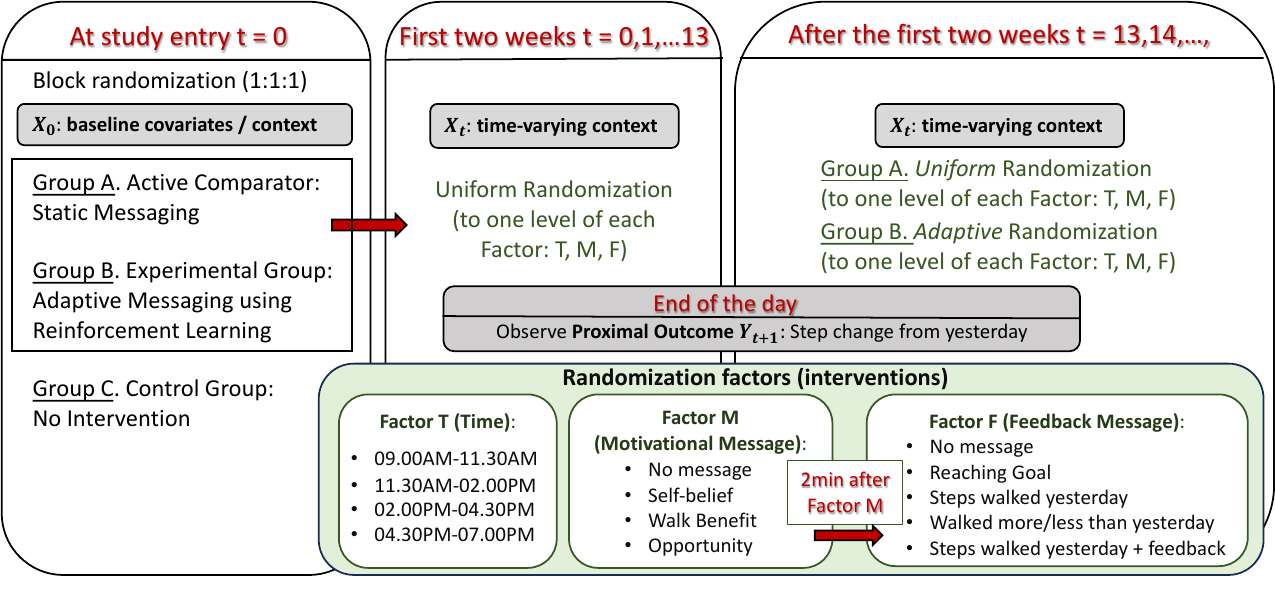}
    \caption{Schematic of the MRT design of the \textit{DIAMANTE} Study.}
    \label{fig: mrt-diamante}
\end{figure*}

As an illustrative example, to better understand the characteristics and value of an MRT, let us now consider the \textit{DIAMANTE} study design of a physical activity trial in Figure~\ref{fig: mrt-diamante}.
In this study, the intervention options (i.e., each unique combination of the factor levels) include whether or not to send a text message, which type of message to deliver, and at which time; the proximal outcome is the change in the number of steps the person walked today from yesterday; and the context is given by a set of user's individual variables such as baseline health status or study day.

At the macro level, to assess the benefits of optimized JITAIs, in addition to evaluating the causal role of each intervention, users are randomized to different study groups (see Figure~\ref{fig: mrt-diamante}), including a static (control) group, a uniform random (nonoptimized) group, and an RL-based (adaptive, optimized) group. In noncontrol groups, users are randomized every day to receive a combination of the different factors' levels, delivered within different time frames. The adaptive RL-based optimized group strategy is illustrated in more detail in Section~\ref{Sec: TS_Diamante}.

\subsection{Case study: DIAMANTE trial} \label{Sec: TS_Diamante}
The \textit{DIAMANTE} trial~\citep{aguilera_mhealth_2020} is an mHealth study that we designed in collaboration with a team of clinicians, phychologists, and computer scientists, among others. The general objective of the study was to encourage users to become more physically active by sending them suitable text messages, and the design involved RL architectures. The overview of the trial is given in Figure~\ref{fig: mrt-diamante}, and preliminary (pilot data) results are presented in~\cite{figueroa_daily_2022}.

During the pilot study, we implemented an adaptive RL mechanisms on a daily basis for deciding on: 1) the feedback message, 2) the motivation message, and 3) the timing of the message. Each combination of levels of the three experimental factors represented an arm or action. To increase personalization, the decision about which message to send also took into consideration contextual variables: time-independent variables such as baseline sociodemographic information and time-dependent covariates, including the day of the week (Monday-Sunday), the data steps of the previous day and the number of days since messages from different categories were sent. For this study, we adopted a MAB strategy adapted to the specific setting of mHealth and the high dimensionality of the context, which we detail below.

\begin{description}
    \item [\textbf{Algorithm}] In this trial, we employed RL only for one of the groups, regarded as the adaptive experimental group. We proposed the contextual linear TS algorithm and decided to implement it after the first two weeks, during which text messages were sent uniformly at random (analogous to an initial ``burn-in'' period, or, more appropriately, an ``internal pilot'' to acquire some prior data to feed into the main algorithm). The choice of TS was motivated by several reasons. First, its empirical and theoretical properties have been well studied and have shown great theoretical and empirical performances \citep{chapelle2011empirical, agrawal2013thompson}. Second, it is computationally efficient, and thus particularly suitable for online learning \citep{russo2017tutorial}. Third, it is a randomized algorithm and, as such, mitigates different forms of biases and allows causal inference \citep{rosenberger2019randomization}. Finally, TS has been widely applied in real-world applications, including mHealth \citep{liao2020personalized} and showed promising results, even with small amounts of data \citep{agrawal2013thompson}. 

    \item[\textbf{Regularization}] We used a Bayesian linear regression setting with a \textit{Normal-Inverse-Gamma} (NIG) prior for the regression coefficients (mean and variance). This choice keeps the coefficients small and minimizes overfitting, providing some form of regularization. A Gaussian distribution is assumed for the reward, i.e.,
\begin{align*}
    Y_t|f(X_t, A_t), \boldsymbol\beta, \sigma^2 \sim \mathcal{N}(f(X_t, A_t)^T\boldsymbol\beta, \sigma^2),\quad t = 1,2,\dots
\end{align*}
with unknown coefficient vector $\boldsymbol\beta$ and unknown variance $\sigma^2>0$. Following a Bayesian framework, the unknown parameters are assumed to be jointly distributed as a multivariate NIG, that is, 
\begin{align*}
 (\boldsymbol{\beta}, \sigma^2)|\boldsymbol\mu_{\beta}, \Sigma_{\beta}, a, b \sim \text{NIG}_{d+1}(\boldsymbol\mu_{\beta}, \Sigma_{\beta}, a, b),
\end{align*}
where $\boldsymbol\mu_{\beta} \in \mathbb{R}^{d+1}$, with $a$, $b \in \mathbb{R}_{>0}$, are fixed and known prior hyper-parameters. When $\Sigma_{\beta}$, the covariance structure between the $\boldsymbol\beta$ vector, is known, the joint prior distribution can be simplified through a hierarchical representation as:
\begin{align*}
    \boldsymbol\beta|\boldsymbol\mu_{\beta}, \Sigma_{\beta}, \sigma^2 &\sim \mathcal{N}_{d+1}(\boldsymbol\mu_{\beta}, \sigma^2 \Sigma_{\beta}),\\
    \sigma^2|a, b &\sim \text{IG}(a,b).
\end{align*}
Note that this also simplifies the sampling process of the TS algorithm. In fact, to perform a posterior sampling from the NIG posterior distribution, it is enough to sample the unknown variance and mean parameters from their updated Inverse-Gamma and Normal posteriors, which we denote with $\text{IG}(a^*, b^*)$ and $\mathcal{N}_{d+1}(\boldsymbol\mu^*, \tilde{\sigma}^2_{(t)} \Sigma^*)$, respectively, where
\begin{align*}
    \boldsymbol{\mu^*} &= \left(\Sigma_\beta^{-1} + f(X_t, A_t)^Tf(X_t, A_t)\right)^{-1}\left(\Sigma_\beta^{-1}\boldsymbol{\mu}_\beta + f(X_t, A_t)^TY_t\right),\\
    \Sigma^* &= \left ( \Sigma^{-1}_\beta + f(X_t, A_t)^Tf(X_t, A_t)\right)^{-1},\\
    a^* &= a+ \frac{n}{2},\\
    b^* &= b + \frac{1}{2}\left(\boldsymbol{\mu}^T_\beta \Sigma^{-1}_\beta\boldsymbol{\mu}_\beta + Y^TY - \boldsymbol{\mu^*}^T\Sigma^{*-1}\boldsymbol{\mu^*} \right),\\
    \tilde{\sigma}^2_{(t)} &\sim \text{IG}(a^*, b^*).
\end{align*}

The resulting vector $\left\{\tilde{\boldsymbol{\beta}}_{(t)}, \tilde{\sigma}^2_{(t)}\right\}_{t=1}^{T}$, with $\tilde{\sigma}^2_{(t)} \sim \text{IG}(a^*, b^*)$ and $\tilde{\boldsymbol{\beta}}_{(t)} \sim \mathcal{N}_{d+1}(\boldsymbol\mu^*, \tilde{\sigma}^2_{(t)} \Sigma^*)$ provides samples from the joint NIG posterior, while $\left\{\tilde{\boldsymbol{\beta}}_{(t)}\right\}_{t=1}^{T}$ and $\left\{\tilde{\sigma}^2_{(t)}\right\}_{t=1}^{T}$ provide samples from the marginal Normal and IG posterior, respectively. Based on the posterior samples, at each iteration $t$, the optimal arm $\tilde{a}_{(t)}$ is the one that maximizes the posterior estimated expected reward, $f(X_t, A_t)^T\tilde{\boldsymbol{\beta}}_{(t)}$, where the posterior nature is reflected in $\tilde{\boldsymbol{\beta}}_{(t)}$.

\item[\textbf{Habituation}] In many real-world scenarios, temporal changes in the reward distribution are an intrinsic characteristic of the problem, and the stationary assumption may be too simplistic. For example, one may expect that the effectiveness of a specific intervention would deteriorate over time or with continued exposure to that intervention. In mHealth, or more generally in behavioral sciences, this phenomenon, known as \textit{habituation}~\citep{epstein2009habituation}, is a recognized pattern. In a text-messaging application such as the DIAMANTE study, sending the same message category repeatedly over time can eventually result in a decreased response or, even worse, disengagement with the program. Consequently, an arm that was optimal for an individual for a certain number of initial days might lose its effectiveness, and the algorithm should quickly adapt to this shift. 
To account for this departure from the stationarity assumption, we followed a \textit{recovery bandit}~\citep{pike-burke_recovering_2019} approach and modeled the outcome of interest (daily step change) as a function of the number of times since each given text-message category was last sent. Consider $K$ different arms (e.g., message categories) and a fixed number of days $T$ in which only one category can be sent. Now, for each text-message category $A_j,j = 1,\dots K$ and day $t = 0,\dots, T$, we denote by $Z_{A_j,t}$ the number of days since category $A_j$ was last played, where $Z_{A_j,t} \in \mathcal{Z} = \{0,\dots,Z_{\text{max}}\}$ for a finite $Z_{\text{max}} \in \mathbb{N}$. More specifically, at day $t+1$, we have that for each $A_j,j = 1,\dots K$,
\begin{align*}
    Z_{A_j,t+1} = \begin{cases} 
0 & \text{if}\ A_{t} = A_{j}\ (\text{or}\  A_{j,t} = 1), \\
\text{min}\{Z_{\text{max}}, Z_{A_j,t}+1\} & \text{if}\ A_{t} \neq A_{j}\ (\text{or}\  A_{j,t} = 0).
\end{cases}
\end{align*}
The variable $A_{j,t}$ is the dummy version of the message category representing whether the category $A_j$ was chosen at time $t$: $A_{j,t} = 1$ means that on day $t$ category $A_j$ was assigned ($A_{t} = A_{j}$), while $A_{j,t} = 0$ means that on day $t$ a category different from $A_j$ was selected ($A_{t} \neq A_{j}$). 

Let now $\overline{\mathbf{Z}}_t \doteq (\overline{Z}_{A_1,t},\dots,\overline{Z}_{A_K,t}) \doteq (Z_{\text{max}} - Z_{A_1,t},\dots,Z_{\text{max}}-Z_{A_K,t})$, for $t = 1,\dots,T$, be the vector of these auxiliary variables, which we name habituation or recovery context. The idea is that, based on the closeness in time of a certain text message category, the reward might be positively or negatively affected. Particularly, with (negative) habituation, we refer to the case where sending the same category consecutively may cause habituation and loss of its potential effect in terms of step change. In this setting, a higher $\overline{Z}_{A_j,t}$ represents a higher degree of habituation at time $t$ related to the category $A_j$, indicating a higher loss in terms of reward if the same message category is sent over time. In line with the behavioral science literature, we hypothesize that in a fixed number of rounds $Z_{\text{max}}$, a specific category will be exempted from habituation if not sent.
Based on this reasoning, at time $t$, we model the daily step change of category $A_j$ as a linear function of this arm, of the time and/or action invariant context $\mathbf{X}_t$ and of the related arm dependent contextual variable $\overline{Z}_{A_j,t}$, i.e.,
\begin{align*}
  \mathbb{E}[Y_{t} | \mathbf{X}_t, A_t = A_j, \overline{Z}_{A_j,t}] 
  &= f(\mathbf{X}_t, A_j, \overline{Z}_{A_j,t})^T\boldsymbol\beta,\quad j = 1,\dots, K. 
\end{align*}
Thus, the expected reward of every arm changes at each round $t$, and this change depends on whether arm $A_j$ was previously played and how many rounds ago. 
    
\item [\textbf{Reward variable}] As in DTRs, choosing the appropriate reward variable is a major issue, and in mHealth it strongly depends on the mobile or wearable instrument. In this study, we used the daily step counts (collected by the pedometer on the participants' personal phones) as the proximal outcome for physical activity. To account for users' baseline walking propensity, we decided to consider the step change from one day to another, which also showed a closer Gaussian shape.
    
    \item [\textbf{Missing data in the reward variable}] To date, there has been a lack of mHealth studies addressing this problem. However, in an online experimental setting, it is particularly relevant, as it can impact subsequent selection of interventions when reward is missing. In this work, we set as missing all the $0$ step counts; this is done in line with the existing literature, which suggests that this outcome is due to technical errors of the device, or simply users' forgetfulness to carry their phones when walking. Then, taking an exploratory approach, we performed multiple imputation of missing data during post-data collection analysis, more specifically as a sensitivity analysis. During data collection, we used the last observation carried forward technique. 

\end{description}

\section{Final Remarks} \label{Sec: Conclusions}

The content of this work summarizes and highlights the increasing potential and interest in RL and AI in general, to improve healthcare and promote healthy behaviors. However, despite remarkable theoretical results and aside from a few examples, fulfilling the vision of precision health and well-being for all is still far from realization. Although the opportunity and potential of AI to make health systems more efficient, sustainable, and equitable is visible, its real-world application to create personalized and effective care remains largely untapped.

In fact, despite remarkable theoretical results, only a few studies applied RL in real life. Moreover, in many cases, applications simply used RL approaches for solving the problem in relatively simplified settings, thus exhibiting a number of shortcomings and practical limitations and posing interesting technical challenges and open problems. The following are a list of nonexhaustive questions that still need to be fully addressed. How can we better understand and interpret the process of an RL algorithm, which often acts in a black box expressed by, for instance, deep neural networks? How can we adequately adapt the RL strategy to complex disease and behavioral scenarios? This requires formalizing appropriate relationships in the RL process, particularly for the reward function, taking into account domain knowledge about each specific setting, its multiple objectives, and the potential presence of nonstationarities or unstructured data. While, for instance, several software packages exist for implementing many of the reviewed algorithms, these are often suitable only under specific cases (e.g., only continuous and positive rewards) and require detailed knowledge about the software. 

In addition, more work remains to be done to test, validate, and change healthcare practices, navigating complex ethical, technical, and human-centered challenges in privacy, transparency, and fairness~\citep{chien_multi-disciplinary_2022}, among others~\citep[see also][]{deliu2023reinforcement}. Finally, precision health is a highly multidisciplinary domain--intersecting behavioral and clinical domains, as well as methodological area--and is still in its infancy, having been formalized as a progeny of precision medicine only in the last decade~\citep{ryan_defining_2021}. The challenge is thus to unravel the potential of AI in precision health in a cohesive and synergistic way, taking into account the methodological details and their practical adoption. This is the approach we followed in this work, providing a detailed survey of the methodological framework of interest and discussing its benefits in the context of precision and digital health with concrete case studies.


\bibliographystyle{apalike}
\bibliography{main.bib}

\begin{thebibliography}{}

\bibitem[Aaltonen et~al., 2020]{aaltonen_pan-cancer_2020}
Aaltonen, L.~A., Abascal, F., Abeshouse, A., {and others}, and {The ICGC/TCGA Pan-Cancer Analysis of Whole Genomes Consortium} (2020).
\newblock Pan-cancer analysis of whole genomes.
\newblock {\em Nature}, 578(7793):82--93.

\bibitem[Agrawal and Prabakaran, 2020]{agrawal_big_2020}
Agrawal, R. and Prabakaran, S. (2020).
\newblock Big data in digital healthcare: lessons learnt and recommendations for general practice.
\newblock {\em Heredity}, 124(4):525--534.

\bibitem[Agrawal and Goyal, 2013]{agrawal2013thompson}
Agrawal, S. and Goyal, N. (2013).
\newblock Thompson {Sampling} for {Contextual} {Bandits} with {Linear} {Payoffs}.
\newblock In {\em Proceedings of the 30th {International} {Conference} on {Machine} {Learning}}, volume 28(3), pages 127--135. PMLR.

\bibitem[Aguilera et~al., 2020]{aguilera_mhealth_2020}
Aguilera, A., Figueroa, C.~A., Hernandez-Ramos, R., Sarkar, U., Cemballi, A., Gomez-Pathak, L., Miramontes, J., Yom-Tov, E., Chakraborty, B., Yan, X., Xu, J., Modiri, A., Aggarwal, J., Williams, J.~J., and Lyles, C.~R. (2020).
\newblock {mHealth} app using machine learning to increase physical activity in diabetes and depression: clinical trial protocol for the {DIAMANTE} {Study}.
\newblock {\em BMJ Open}, 10(8):e034723.

\bibitem[{All of Us Research Program Investigators} et~al., 2019]{all_of_us_research_program_investigators_all_2019}
{All of Us Research Program Investigators}, Denny, J.~C., Rutter, J.~L., Goldstein, D.~B., Philippakis, A., Smoller, J.~W., Jenkins, G., and Dishman, E. (2019).
\newblock The "{All} of {Us}" {Research} {Program}.
\newblock {\em The New England Journal of Medicine}, 381(7):668--676.

\bibitem[Allen et~al., 2012]{allen_uk_2012}
Allen, N., Sudlow, C., Downey, P., Peakman, T., Danesh, J., Elliott, P., Gallacher, J., Green, J., Matthews, P., Pell, J., Sprosen, T., and Collins, R. (2012).
\newblock {UK} {Biobank}: {Current} status and what it means for epidemiology.
\newblock {\em Health Policy and Technology}, 1(3):123--126.

\bibitem[Almirall et~al., 2014]{almirall_introduction_2014}
Almirall, D., Nahum-Shani, I., Sherwood, N.~E., and Murphy, S.~A. (2014).
\newblock Introduction to {SMART} designs for the development of adaptive interventions: with application to weight loss research.
\newblock {\em Translational Behavioral Medicine}, 4(3):260--274.

\bibitem[Aswani et~al., 2019]{aswani_behavioral_2019}
Aswani, A., Kaminsky, P., Mintz, Y., Flowers, E., and Fukuoka, Y. (2019).
\newblock Behavioral {Modeling} in {Weight} {Loss} {Interventions}.
\newblock {\em European Journal of Operational Research}, 272(3):1058--1072.

\bibitem[Atan et~al., 2018]{atan2018deep}
Atan, O., Jordon, J., and Van Der~Schaar, M. (2018).
\newblock Deep-{Treat}: {Learning} {Optimal} {Personalized} {Treatments} {From} {Observational} {Data} {Using} {Neural} {Networks}.
\newblock {\em Proceedings of the AAAI Conference on Artificial Intelligence}, 32(1).

\bibitem[Auer et~al., 2002a]{auer2002finite}
Auer, P., Cesa-Bianchi, N., and Fischer, P. (2002a).
\newblock Finite-time {Analysis} of the {Multiarmed} {Bandit} {Problem}.
\newblock {\em Machine Learning}, 47(2):235--256.

\bibitem[Auer et~al., 2002b]{auer2002nonstochastic}
Auer, P., Cesa-Bianchi, N., Freund, Y., and Schapire, R.~E. (2002b).
\newblock The {Nonstochastic} {Multiarmed} {Bandit} {Problem}.
\newblock {\em SIAM Journal on Computing}, 32(1):48--77.

\bibitem[Beaglehole et~al., 2011]{beaglehole_priority_2011}
Beaglehole, R., Bonita, R., Horton, R., Adams, C., Alleyne, G., Asaria, P., Baugh, V., Bekedam, H., Billo, N., Casswell, S., Cecchini, M., Colagiuri, R., Colagiuri, S., Collins, T., Ebrahim, S., Engelgau, M., Galea, G., Gaziano, T., Geneau, R., Haines, A., Hospedales, J., Jha, P., Keeling, A., Leeder, S., Lincoln, P., McKee, M., Mackay, J., Magnusson, R., Moodie, R., Mwatsama, M., Nishtar, S., Norrving, B., Patterson, D., Piot, P., Ralston, J., Rani, M., Reddy, K.~S., Sassi, F., Sheron, N., Stuckler, D., Suh, I., Torode, J., Varghese, C., and Watt, J. (2011).
\newblock Priority actions for the non-communicable disease crisis.
\newblock {\em The Lancet}, 377(9775):1438--1447.

\bibitem[Bell et~al., 2020]{bell_notifications_2020}
Bell, L., Garnett, C., Qian, T., Perski, O., Potts, H. W.~W., and Williamson, E. (2020).
\newblock Notifications to {Improve} {Engagement} {With} an {Alcohol} {Reduction} {App}: {Protocol} for a {Micro}-{Randomized} {Trial}.
\newblock {\em JMIR research protocols}, 9(8):e18690.

\bibitem[Bellman, 1957]{Bellman1957}
Bellman, R. (1957).
\newblock {\em Dynamic programming}.
\newblock Dover Publications, Mineola, N.Y.

\bibitem[Bertsekas, 2019]{bertsekas2019reinforcement}
Bertsekas, D.~P. (2019).
\newblock {\em Reinforcement {Learning} and {Optimal} {Control}}.
\newblock Athena Scientific, Belmont, Massachusetts, 2 edition.

\bibitem[Bishop, 2006]{bishop_pattern_2006}
Bishop, C.~M. (2006).
\newblock {\em Pattern recognition and machine learning}.
\newblock Information science and statistics. Springer, New York.

\bibitem[Bubeck and Cesa-Bianchi, 2012]{bubeck2012regret}
Bubeck, S. and Cesa-Bianchi, N. (2012).
\newblock Regret {Analysis} of {Stochastic} and {Nonstochastic} {Multi}-armed {Bandit} {Problems}.
\newblock {\em Foundations and Trends® in Machine Learning}, 5(1):1--122.

\bibitem[Camacho et~al., 2018]{camacho_next-generation_2018}
Camacho, D.~M., Collins, K.~M., Powers, R.~K., Costello, J.~C., and Collins, J.~J. (2018).
\newblock Next-{Generation} {Machine} {Learning} for {Biological} {Networks}.
\newblock {\em Cell}, 173(7):1581--1592.

\bibitem[Chakraborty, 2009]{chakraborty_study_2009}
Chakraborty, B. (2009).
\newblock {\em A {Study} of {Non}-regularity in {Dynamic} {Treatment} {Regimes} and {Some} {Design} {Considerations} for {Multicomponent} {Interventions}.}
\newblock {PhD} {Thesis}, University of Michigan.

\bibitem[Chakraborty et~al., 2013]{chakraborty2013inference}
Chakraborty, B., Laber, E.~B., and Zhao, Y. (2013).
\newblock Inference for {Optimal} {Dynamic} {Treatment} {Regimes} {Using} an {Adaptive} \textit{m}‐out‐of‐\textit{n} {Bootstrap} {Scheme}.
\newblock {\em Biometrics}, 69(3):714--723.

\bibitem[Chakraborty and Moodie, 2013]{chakraborty_statistical_2013}
Chakraborty, B. and Moodie, E. E.~M. (2013).
\newblock {\em Statistical {Methods} for {Dynamic} {Treatment} {Regimes}: {Reinforcement} {Learning}, {Causal} {Inference}, and {Personalized} {Medicine}}.
\newblock Statistics for {Biology} and {Health}. Springer, New York, NY.

\bibitem[Chakraborty et~al., 2010]{chakraborty2010inference}
Chakraborty, B., Murphy, S., and Strecher, V. (2010).
\newblock Inference for non-regular parameters in optimal dynamic treatment regimes.
\newblock {\em Statistical Methods in Medical Research}, 19(3):317--343.

\bibitem[Chakraborty and Murphy, 2014]{chakraborty_dynamic_2014}
Chakraborty, B. and Murphy, S.~A. (2014).
\newblock Dynamic {Treatment} {Regimes}.
\newblock {\em Annual Review of Statistics and Its Application}, 1(1):447--464.

\bibitem[Chapelle and Li, 2011]{chapelle2011empirical}
Chapelle, O. and Li, L. (2011).
\newblock An {Empirical} {Evaluation} of {Thompson} {Sampling}.
\newblock In {\em Proceedings of the 24th {International} {Conference} on {Neural} {Information} {Processing} {Systems}}, volume~24, pages 2249--2257.

\bibitem[Chen et~al., 2016]{chen_personalized_2016}
Chen, G., Zeng, D., and Kosorok, M.~R. (2016).
\newblock Personalized {Dose} {Finding} {Using} {Outcome} {Weighted} {Learning}.
\newblock {\em Journal of the American Statistical Association}, 111(516):1509--1521.

\bibitem[Chen et~al., 2018]{chen_estimating_2018}
Chen, J., Fu, H., He, X., Kosorok, M.~R., and Liu, Y. (2018).
\newblock Estimating individualized treatment rules for ordinal treatments.
\newblock {\em Biometrics}, 74(3):924--933.

\bibitem[Chien et~al., 2022]{chien_multi-disciplinary_2022}
Chien, I., Deliu, N., Turner, R., Weller, A., Villar, S., and Kilbertus, N. (2022).
\newblock Multi-disciplinary fairness considerations in machine learning for clinical trials.
\newblock In {\em 2022 {ACM} {Conference} on {Fairness}, {Accountability}, and {Transparency}}, pages 906--924, Seoul Republic of Korea. ACM.

\bibitem[Chu et~al., 2011]{chu2011contextual}
Chu, W., Li, L., Reyzin, L., and Schapire, R. (2011).
\newblock Contextual {Bandits} with {Linear} {Payoff} {Functions}.
\newblock In {\em Proceedings of the {Fourteenth} {International} {Conference} on {Artificial} {Intelligence} and {Statistics}}, pages 208--214. JMLR Workshop and Conference Proceedings.

\bibitem[Collins and Varmus, 2015]{collins_new_2015}
Collins, F.~S. and Varmus, H. (2015).
\newblock A {New} {Initiative} on {Precision} {Medicine}.
\newblock {\em New England Journal of Medicine}, 372(9):793--795.

\bibitem[Collins et~al., 2009]{collins_comparison_2009}
Collins, L.~M., Chakraborty, B., Murphy, S.~A., and Strecher, V. (2009).
\newblock Comparison of a phased experimental approach and a single randomized clinical trial for developing multicomponent behavioral interventions.
\newblock {\em Clinical Trials}, 6(1):5--15.

\bibitem[Collins et~al., 2004]{collins_conceptual_2004}
Collins, L.~M., Murphy, S.~A., and Bierman, K.~L. (2004).
\newblock A {Conceptual} {Framework} for {Adaptive} {Preventive} {Interventions}.
\newblock {\em Prevention Science}, 5(3):185--196.

\bibitem[De~Lara et~al., 2008]{de_lara_sustainable_2008}
De~Lara, M., De~Lara, M., and Doyen, L. (2008).
\newblock {\em Sustainable management of natural resources: mathematical models and methods}.
\newblock Environmental science and engineering {Subseries} {Environmental} science. Springer, Berlin Heidelberg.

\bibitem[Deliu and Chakraborty, 2022]{deliu_dynamic_2022}
Deliu, N. and Chakraborty, B. (2022).
\newblock Dynamic {Treatment} {Regimes} for {Optimizing} {Healthcare}.
\newblock In Chen, X., Jasin, S., and Shi, C., editors, {\em The {Elements} of {Joint} {Learning} and {Optimization} in {Operations} {Management}}, Springer {Series} in {Supply} {Chain} {Management}, pages 391--444. Springer International Publishing, Cham.

\bibitem[Deliu et~al., 2024]{deliu2023reinforcement}
Deliu, N., Williams, J.~J., and Chakraborty, B. (2024).
\newblock Reinforcement {Learning} in {Modern} {Biostatistics}: {Constructing} {Optimal} {Adaptive} {Interventions}.
\newblock {\em International Statistical Review}, In Press.

\bibitem[Deo, 2015]{deo_machine_2015}
Deo, R.~C. (2015).
\newblock Machine {Learning} in {Medicine}.
\newblock {\em Circulation}, 132(20):1920--1930.

\bibitem[Epstein et~al., 2009]{epstein2009habituation}
Epstein, L.~H., Temple, J.~L., Roemmich, J.~N., and Bouton, M.~E. (2009).
\newblock Habituation as a determinant of human food intake.
\newblock {\em Psychological Review}, 116(2):384--407.

\bibitem[Ertefaie and Strawderman, 2018]{ertefaie2018constructing}
Ertefaie, A. and Strawderman, R.~L. (2018).
\newblock Constructing dynamic treatment regimes over indefinite time horizons.
\newblock {\em Biometrika}, 105(4):963--977.

\bibitem[Figueroa et~al., 2022]{figueroa_daily_2022}
Figueroa, C.~A., Deliu, N., Chakraborty, B., Modiri, A., Xu, J., Aggarwal, J., Jay~Williams, J., Lyles, C., and Aguilera, A. (2022).
\newblock Daily {Motivational} {Text} {Messages} to {Promote} {Physical} {Activity} in {University} {Students}: {Results} {From} a {Microrandomized} {Trial}.
\newblock {\em Annals of Behavioral Medicine}, 56(2):212--218.

\bibitem[Forman et~al., 2019]{forman2019can}
Forman, E.~M., Kerrigan, S.~G., Butryn, M.~L., Juarascio, A.~S., Manasse, S.~M., Ontañón, S., Dallal, D.~H., Crochiere, R.~J., and Moskow, D. (2019).
\newblock Can the artificial intelligence technique of reinforcement learning use continuously-monitored digital data to optimize treatment for weight loss?
\newblock {\em Journal of Behavioral Medicine}, 42(2):276--290.

\bibitem[Fu et~al., 2019]{fu_robust_2019}
Fu, S., He, Q., Zhang, S., and Liu, Y. (2019).
\newblock Robust outcome weighted learning for optimal individualized treatment rules.
\newblock {\em Journal of Biopharmaceutical Statistics}, 29(4):606--624.

\bibitem[Gambhir et~al., 2018]{gambhir_toward_2018}
Gambhir, S.~S., Ge, T.~J., Vermesh, O., and Spitler, R. (2018).
\newblock Toward achieving precision health.
\newblock {\em Science Translational Medicine}, 10(430):eaao3612.

\bibitem[Garnett et~al., 2019]{garnett2019development}
Garnett, C., Crane, D., West, R., Brown, J., and Michie, S. (2019).
\newblock The development of \textit{{Drink} {Less}} : an alcohol reduction smartphone app for excessive drinkers.
\newblock {\em Translational Behavioral Medicine}, 9(2):296--307.

\bibitem[Goldberg and Kosorok, 2012]{goldberg_q-learning_2012}
Goldberg, Y. and Kosorok, M.~R. (2012).
\newblock Q-learning with censored data.
\newblock {\em The Annals of Statistics}, 40(1).

\bibitem[Goldstein et~al., 2017]{goldstein2017return}
Goldstein, S.~P., Evans, B.~C., Flack, D., Juarascio, A., Manasse, S., Zhang, F., and Forman, E.~M. (2017).
\newblock Return of the {JITAI}: {Applying} a {Just}-in-{Time} {Adaptive} {Intervention} {Framework} to the {Development} of m-{Health} {Solutions} for {Addictive} {Behaviors}.
\newblock {\em International Journal of Behavioral Medicine}, 24(5):673--682.

\bibitem[Goodfellow et~al., 2016]{goodfellow2016deep}
Goodfellow, I., Bengio, Y., and Courville, A. (2016).
\newblock {\em Deep learning}.
\newblock Adaptive {Computation} and {Machine} {Learning}. The MIT Press, Cambridge, Massachusetts.

\bibitem[Gottesman et~al., 2019]{gottesman_guidelines_2019}
Gottesman, O., Johansson, F., Komorowski, M., Faisal, A., Sontag, D., Doshi-Velez, F., and Celi, L.~A. (2019).
\newblock Guidelines for reinforcement learning in healthcare.
\newblock {\em Nature Medicine}, 25(1):16--18.

\bibitem[Greenewald et~al., 2017]{greenewald17}
Greenewald, K., Tewari, A., Klasnja, P., and Murphy, S. (2017).
\newblock Action centered contextual bandits.
\newblock In {\em Proceedings of the 31st {International} {Conference} on {Neural} {Information} {Processing} {Systems}}, {NIPS}'17, pages 5979--5987, Red Hook, NY, USA. Curran Associates Inc.

\bibitem[Grondman et~al., 2012]{grondman_survey_2012}
Grondman, I., Busoniu, L., Lopes, G. A.~D., and Babuska, R. (2012).
\newblock A {Survey} of {Actor}-{Critic} {Reinforcement} {Learning}: {Standard} and {Natural} {Policy} {Gradients}.
\newblock {\em IEEE Transactions on Systems, Man, and Cybernetics, Part C (Applications and Reviews)}, 42(6):1291--1307.

\bibitem[Guan et~al., 2021]{guan2019direct}
Guan, Y., Li, S.~E., Duan, J., Li, J., Ren, Y., Sun, Q., and Cheng, B. (2021).
\newblock Direct and indirect reinforcement learning.
\newblock {\em International Journal of Intelligent Systems}, 36(8):4439--4467.

\bibitem[Hardeman et~al., 2019]{hardeman2019systematic}
Hardeman, W., Houghton, J., Lane, K., Jones, A., and Naughton, F. (2019).
\newblock A systematic review of just-in-time adaptive interventions ({JITAIs}) to promote physical activity.
\newblock {\em International Journal of Behavioral Nutrition and Physical Activity}, 16(1):31.

\bibitem[Hastie et~al., 2009]{hastie_elements_2009}
Hastie, T., Tibshirani, R., and Friedman, J. (2009).
\newblock {\em The {Elements} of {Statistical} {Learning}}.
\newblock Springer {Series} in {Statistics}. Springer New York, New York, NY.

\bibitem[Hernan and Robins, 2023]{hernan_causal_2023}
Hernan, M.~A. and Robins, J.~M. (2023).
\newblock {\em Causal {Inference}: {What} {If}}.
\newblock CRC Press, Boca Raton.

\bibitem[Istepanian et~al., 2006]{istepanian2007m}
Istepanian, R. S.~H., Laxminarayan, S., and Pattichis, C.~S., editors (2006).
\newblock {\em M-{Health}: {Emerging} {Mobile} {Health} {Systems}}.
\newblock Topics in {Biomedical} {Engineering}. {International} {Book} {Series} ({ITBE}). Springer, New York, N.Y.

\bibitem[Johnson et~al., 2021]{johnson_precision_2021}
Johnson, K.~B., Wei, W., Weeraratne, D., Frisse, M.~E., Misulis, K., Rhee, K., Zhao, J., and Snowdon, J.~L. (2021).
\newblock Precision {Medicine}, {AI}, and the {Future} of {Personalized} {Health} {Care}.
\newblock {\em Clinical and Translational Science}, 14(1):86--93.

\bibitem[Kitagawa and Tetenov, 2018]{kitagawa_who_2018}
Kitagawa, T. and Tetenov, A. (2018).
\newblock Who {Should} {Be} {Treated}? {Empirical} {Welfare} {Maximization} {Methods} for {Treatment} {Choice}.
\newblock {\em Econometrica}, 86(2):591--616.

\bibitem[Klasnja et~al., 2015]{klasnja2015microrandomized}
Klasnja, P., Hekler, E.~B., Shiffman, S., Boruvka, A., Almirall, D., Tewari, A., and Murphy, S.~A. (2015).
\newblock Microrandomized trials: {An} experimental design for developing just-in-time adaptive interventions.
\newblock {\em Health Psychology}, 34(Suppl):1220--1228.

\bibitem[Kosorok and Laber, 2019]{kosorok_precision_2019}
Kosorok, M.~R. and Laber, E.~B. (2019).
\newblock Precision {Medicine}.
\newblock {\em Annual Review of Statistics and Its Application}, 6(1):263--286.

\bibitem[Kourou et~al., 2015]{kourou_machine_2015}
Kourou, K., Exarchos, T.~P., Exarchos, K.~P., Karamouzis, M.~V., and Fotiadis, D.~I. (2015).
\newblock Machine learning applications in cancer prognosis and prediction.
\newblock {\em Computational and Structural Biotechnology Journal}, 13:8--17.

\bibitem[Laber et~al., 2014]{laber_set-valued_2014}
Laber, E.~B., Lizotte, D.~J., and Ferguson, B. (2014).
\newblock Set-valued dynamic treatment regimes for competing outcomes.
\newblock {\em Biometrics}, 70(1):53--61.

\bibitem[Laber and Zhao, 2015]{laber_tree-based_2015}
Laber, E.~B. and Zhao, Y.~Q. (2015).
\newblock Tree-based methods for individualized treatment regimes.
\newblock {\em Biometrika}, 102(3):501--514.

\bibitem[Lavori and Dawson, 2000]{lavori_design_2000}
Lavori, P.~W. and Dawson, R. (2000).
\newblock A design for testing clinical strategies: biased adaptive within‐subject randomization.
\newblock {\em Journal of the Royal Statistical Society: Series A (Statistics in Society)}, 163(1):29--38.

\bibitem[Lavori and Dawson, 2004]{lavori_dynamic_2004}
Lavori, P.~W. and Dawson, R. (2004).
\newblock Dynamic treatment regimes: practical design considerations.
\newblock {\em Clinical Trials}, 1(1):9--20.

\bibitem[Lei, 2016]{lei2016online}
Lei, H. (2016).
\newblock {\em An {Online} {Actor} {Critic} {Algorithm} and a {Statistical} {Decision} {Procedure} for {Personalizing} {Intervention}}.
\newblock {PhD} {Thesis}, University of Michigan.

\bibitem[Li et~al., 2010]{li2010contextual}
Li, L., Chu, W., Langford, J., and Schapire, R.~E. (2010).
\newblock A contextual-bandit approach to personalized news article recommendation.
\newblock In {\em Proceedings of the 19th international conference on {World} wide web}, pages 661--670, Raleigh North Carolina USA. ACM.

\bibitem[Liao et~al., 2020]{liao2020personalized}
Liao, P., Greenewald, K., Klasnja, P., and Murphy, S. (2020).
\newblock Personalized {HeartSteps}: {A} {Reinforcement} {Learning} {Algorithm} for {Optimizing} {Physical} {Activity}.
\newblock {\em Proceedings of the ACM on Interactive, Mobile, Wearable and Ubiquitous Technologies}, 4(1):1--22.

\bibitem[Liu et~al., 2023]{liu_microrandomized_2023}
Liu, X., Deliu, N., and Chakraborty, B. (2023).
\newblock Microrandomized {Trials}: {Developing} {Just}-in-{Time} {Adaptive} {Interventions} for {Better} {Public} {Health}.
\newblock {\em American Journal of Public Health}, 113(1):60--69.

\bibitem[Liu et~al., 2017]{liu2017deep}
Liu, Y., Logan, B., Liu, N., Xu, Z., Tang, J., and Wang, Y. (2017).
\newblock Deep {Reinforcement} {Learning} for {Dynamic} {Treatment} {Regimes} on {Medical} {Registry} {Data}.
\newblock In {\em 2017 {IEEE} {International} {Conference} on {Healthcare} {Informatics} ({ICHI})}, pages 380--385, Park City, UT. IEEE.

\bibitem[Liu et~al., 2018]{liu2018augmented}
Liu, Y., Wang, Y., Kosorok, M.~R., Zhao, Y., and Zeng, D. (2018).
\newblock Augmented outcome-weighted learning for estimating optimal dynamic treatment regimens: {Augmented} {Outcome}-weighted {Learning}.
\newblock {\em Statistics in Medicine}, 37(26):3776--3788.

\bibitem[Luckett et~al., 2020]{luckett2020estimating}
Luckett, D.~J., Laber, E.~B., Kahkoska, A.~R., Maahs, D.~M., Mayer-Davis, E., and Kosorok, M.~R. (2020).
\newblock Estimating {Dynamic} {Treatment} {Regimes} in {Mobile} {Health} {Using} {V}-{Learning}.
\newblock {\em Journal of the American Statistical Association}, 115(530):692--706.

\bibitem[Luedtke and van~der Laan, 2016]{luedtke_super-learning_2016}
Luedtke, A.~R. and van~der Laan, M.~J. (2016).
\newblock Super-{Learning} of an {Optimal} {Dynamic} {Treatment} {Rule}.
\newblock {\em The international journal of biostatistics}, 12(1):305--332.

\bibitem[MacKinnon et~al., 2007]{mackinnon_mediation_2007}
MacKinnon, D.~P., Fairchild, A.~J., and Fritz, M.~S. (2007).
\newblock Mediation {Analysis}.
\newblock {\em Annual Review of Psychology}, 58(1):593--614.

\bibitem[Mahar et~al., 2021]{mahar_scoping_2021}
Mahar, R.~K., McGuinness, M.~B., Chakraborty, B., Carlin, J.~B., IJzerman, M.~J., and Simpson, J.~A. (2021).
\newblock A scoping review of studies using observational data to optimise dynamic treatment regimens.
\newblock {\em BMC Medical Research Methodology}, 21(1):39.

\bibitem[Marsh and Cormier, 2002]{marsh2001spline}
Marsh, L. and Cormier, D. (2002).
\newblock {\em Spline {Regression} {Models}}.
\newblock SAGE Publications, Inc., 2455 Teller Road, Thousand Oaks California 91320 United States of America.

\bibitem[McKinney et~al., 2020]{mckinney_international_2020}
McKinney, S.~M., Sieniek, M., Godbole, V., Godwin, J., Antropova, N., Ashrafian, H., Back, T., Chesus, M., Corrado, G.~S., Darzi, A., Etemadi, M., Garcia-Vicente, F., Gilbert, F.~J., Halling-Brown, M., Hassabis, D., Jansen, S., Karthikesalingam, A., Kelly, C.~J., King, D., Ledsam, J.~R., Melnick, D., Mostofi, H., Peng, L., Reicher, J.~J., Romera-Paredes, B., Sidebottom, R., Suleyman, M., Tse, D., Young, K.~C., De~Fauw, J., and Shetty, S. (2020).
\newblock International evaluation of an {AI} system for breast cancer screening.
\newblock {\em Nature}, 577(7788):89--94.

\bibitem[Mnih et~al., 2015]{mnih2015human}
Mnih, V., Kavukcuoglu, K., Silver, D., Rusu, A.~A., Veness, J., Bellemare, M.~G., Graves, A., Riedmiller, M., Fidjeland, A.~K., Ostrovski, G., Petersen, S., Beattie, C., Sadik, A., Antonoglou, I., King, H., Kumaran, D., Wierstra, D., Legg, S., and Hassabis, D. (2015).
\newblock Human-level control through deep reinforcement learning.
\newblock {\em Nature}, 518(7540):529--533.

\bibitem[Montoya et~al., 2023]{montoya_optimal_2023}
Montoya, L.~M., van~der Laan, M.~J., Luedtke, A.~R., Skeem, J.~L., Coyle, J.~R., and Petersen, M.~L. (2023).
\newblock The optimal dynamic treatment rule superlearner: considerations, performance, and application to criminal justice interventions.
\newblock {\em The International Journal of Biostatistics}, 19(1):217--238.

\bibitem[Moodie et~al., 2014]{moodie_q-learning_2014}
Moodie, E. E.~M., Dean, N., and Sun, Y.~R. (2014).
\newblock Q-{Learning}: {Flexible} {Learning} {About} {Useful} {Utilities}.
\newblock {\em Statistics in Biosciences}, 6(2):223--243.

\bibitem[Murphy, 2003]{murphy_optimal_2003}
Murphy, S.~A. (2003).
\newblock Optimal dynamic treatment regimes.
\newblock {\em Journal of the Royal Statistical Society: Series B (Statistical Methodology)}, 65(2):331--355.

\bibitem[Murphy, 2005a]{murphy_experimental_2005}
Murphy, S.~A. (2005a).
\newblock An experimental design for the development of adaptive treatment strategies.
\newblock {\em Statistics in Medicine}, 24(10):1455--1481.

\bibitem[Murphy, 2005b]{murphy2005generalization}
Murphy, S.~A. (2005b).
\newblock A {Generalization} {Error} for {Q}-{Learning}.
\newblock {\em The Journal of Machine Learning Research}, 6:1073--1097.

\bibitem[Myszczynska et~al., 2020]{myszczynska_applications_2020}
Myszczynska, M.~A., Ojamies, P.~N., Lacoste, A. M.~B., Neil, D., Saffari, A., Mead, R., Hautbergue, G.~M., Holbrook, J.~D., and Ferraiuolo, L. (2020).
\newblock Applications of machine learning to diagnosis and treatment of neurodegenerative diseases.
\newblock {\em Nature Reviews. Neurology}, 16(8):440--456.

\bibitem[Nahum-Shani and Almirall, 2019]{nahum-shani_introduction_2019}
Nahum-Shani, I. and Almirall, D. (2019).
\newblock An {Introduction} to {Adaptive} {Interventions} and {SMART} {Designs} in {Education} ({NCSER} 2020-001).
\newblock Technical report, U.S. Department of Education. Washington, DC: National Center for Special Education Research.

\bibitem[Nahum-Shani et~al., 2015]{nahum2015building}
Nahum-Shani, I., Hekler, E.~B., and Spruijt-Metz, D. (2015).
\newblock Building health behavior models to guide the development of just-in-time adaptive interventions: {A} pragmatic framework.
\newblock {\em Health Psychology}, 34(Suppl):1209--1219.

\bibitem[Nahum-Shani et~al., 2018]{nahum2018just}
Nahum-Shani, I., Smith, S.~N., Spring, B.~J., Collins, L.~M., Witkiewitz, K., Tewari, A., and Murphy, S.~A. (2018).
\newblock Just-in-{Time} {Adaptive} {Interventions} ({JITAIs}) in {Mobile} {Health}: {Key} {Components} and {Design} {Principles} for {Ongoing} {Health} {Behavior} {Support}.
\newblock {\em Annals of Behavioral Medicine}, 52(6):446--462.

\bibitem[Naughton, 2017]{naughton2017delivering}
Naughton, F. (2017).
\newblock Delivering "{Just}-{In}-{Time}" {Smoking} {Cessation} {Support} {Via} {Mobile} {Phones}: {Current} {Knowledge} and {Future} {Directions}.
\newblock {\em Nicotine \& Tobacco Research: Official Journal of the Society for Research on Nicotine and Tobacco}, 19(3):379--383.

\bibitem[Navarro-Barrientos et~al., 2011]{navarro-barrientos_dynamical_2011}
Navarro-Barrientos, J.-E., Rivera, D.~E., and Collins, L.~M. (2011).
\newblock A dynamical model for describing behavioural interventions for weight loss and body composition change.
\newblock {\em Mathematical and computer modelling of dynamical systems}, 17(2):183--203.

\bibitem[Ng and Russell, 2000]{ng2000algorithms}
Ng, A.~Y. and Russell, S.~J. (2000).
\newblock Algorithms for {Inverse} {Reinforcement} {Learning}.
\newblock In {\em Proceedings of the {Seventeenth} {International} {Conference} on {Machine} {Learning}}, {ICML} '00, pages 663--670, San Francisco, CA, USA. Morgan Kaufmann Publishers Inc.

\bibitem[Nie et~al., 2021]{nie_learning_2021}
Nie, X., Brunskill, E., and Wager, S. (2021).
\newblock Learning {When}-to-{Treat} {Policies}.
\newblock {\em Journal of the American Statistical Association}, 116(533):392--409.

\bibitem[Pfammatter et~al., 2019]{pfammatter_smart_2019}
Pfammatter, A.~F., Nahum-Shani, I., DeZelar, M., Scanlan, L., McFadden, H.~G., Siddique, J., Hedeker, D., and Spring, B. (2019).
\newblock {SMART}: {Study} protocol for a sequential multiple assignment randomized controlled trial to optimize weight loss management.
\newblock {\em Contemporary Clinical Trials}, 82:36--45.

\bibitem[Pike-Burke and Grünewälder, 2019]{pike-burke_recovering_2019}
Pike-Burke, C. and Grünewälder, S. (2019).
\newblock Recovering bandits.
\newblock In {\em Proceedings of the 33rd {International} {Conference} on {Neural} {Information} {Processing} {Systems}}, 1265, pages 14122--14131. Curran Associates Inc., Red Hook, NY, USA.

\bibitem[Prince et~al., 2015]{prince_burden_2015}
Prince, M.~J., Wu, F., Guo, Y., Gutierrez~Robledo, L.~M., O'Donnell, M., Sullivan, R., and Yusuf, S. (2015).
\newblock The burden of disease in older people and implications for health policy and practice.
\newblock {\em The Lancet}, 385(9967):549--562.

\bibitem[{PsychENCODE Consortium} et~al., 2015]{psychencode_consortium_psychencode_2015}
{PsychENCODE Consortium}, Akbarian, S., Liu, C., Knowles, J.~A., Vaccarino, F.~M., Farnham, P.~J., Crawford, G.~E., Jaffe, A.~E., Pinto, D., Dracheva, S., Geschwind, D.~H., Mill, J., Nairn, A.~C., Abyzov, A., Pochareddy, S., Prabhakar, S., Weissman, S., Sullivan, P.~F., State, M.~W., Weng, Z., Peters, M.~A., White, K.~P., Gerstein, M.~B., Amiri, A., Armoskus, C., Ashley-Koch, A.~E., Bae, T., Beckel-Mitchener, A., Berman, B.~P., Coetzee, G.~A., Coppola, G., Francoeur, N., Fromer, M., Gao, R., Grennan, K., Herstein, J., Kavanagh, D.~H., Ivanov, N.~A., Jiang, Y., Kitchen, R.~R., Kozlenkov, A., Kundakovic, M., Li, M., Li, Z., Liu, S., Mangravite, L.~M., Mattei, E., Markenscoff-Papadimitriou, E., Navarro, F. C.~P., North, N., Omberg, L., Panchision, D., Parikshak, N., Poschmann, J., Price, A.~J., Purcaro, M., Reddy, T.~E., Roussos, P., Schreiner, S., Scuderi, S., Sebra, R., Shibata, M., Shieh, A.~W., Skarica, M., Sun, W., Swarup, V., Thomas, A., Tsuji, J., van Bakel, H., Wang, D., Wang, Y., Wang, K., Werling,
  D.~M., Willsey, A.~J., Witt, H., Won, H., Wong, C. C.~Y., Wray, G.~A., Wu, E.~Y., Xu, X., Yao, L., Senthil, G., Lehner, T., Sklar, P., and Sestan, N. (2015).
\newblock The {PsychENCODE} project.
\newblock {\em Nature Neuroscience}, 18(12):1707--1712.

\bibitem[Qian and Murphy, 2011]{qian_performance_2011}
Qian, M. and Murphy, S.~A. (2011).
\newblock Performance guarantees for individualized treatment rules.
\newblock {\em The Annals of Statistics}, 39(2):1180--1210.

\bibitem[Raghu et~al., 2017]{raghu_continuous_2017}
Raghu, A., Komorowski, M., Celi, L.~A., Szolovits, P., and Ghassemi, M. (2017).
\newblock Continuous {State}-{Space} {Models} for {Optimal} {Sepsis} {Treatment}: a {Deep} {Reinforcement} {Learning} {Approach}.
\newblock In {\em Proceedings of the 2nd {Machine} {Learning} for {Healthcare} {Conference}}, pages 147--163. PMLR.

\bibitem[Rajkomar et~al., 2019]{rajkomar_machine_2019}
Rajkomar, A., Dean, J., and Kohane, I. (2019).
\newblock Machine {Learning} in {Medicine}.
\newblock {\em New England Journal of Medicine}, 380(14):1347--1358.

\bibitem[Rivera et~al., 2007]{rivera_using_2007}
Rivera, D.~E., Pew, M.~D., and Collins, L.~M. (2007).
\newblock Using {Engineering} {Control} {Principles} to {Inform} the {Design} of {Adaptive} {Interventions}: {A} {Conceptual} {Introduction}.
\newblock {\em Drug and alcohol dependence}, 88(Suppl 2):S31--S40.

\bibitem[Robins, 1994]{robins1994correcting}
Robins, J.~M. (1994).
\newblock Correcting for non-compliance in randomized trials using structural nested mean models.
\newblock {\em Communications in Statistics - Theory and Methods}, 23(8):2379--2412.

\bibitem[Robins, 2000]{robins2000marginal}
Robins, J.~M. (2000).
\newblock Marginal {Structural} {Models} versus {Structural} nested {Models} as {Tools} for {Causal} inference.
\newblock In Halloran, M.~E. and Berry, D., editors, {\em Statistical {Models} in {Epidemiology}, the {Environment}, and {Clinical} {Trials}}, The {IMA} {Volumes} in {Mathematics} and its {Applications}, pages 95--133, New York, NY. Springer.

\bibitem[Robins, 2004]{robins2004optimal}
Robins, J.~M. (2004).
\newblock Optimal {Structural} {Nested} {Models} for {Optimal} {Sequential} {Decisions}.
\newblock In Lin, D.~Y. and Heagerty, P.~J., editors, {\em Proceedings of the {Second} {Seattle} {Symposium} in {Biostatistics}: {Analysis} of {Correlated} {Data}}, Lecture {Notes} in {Statistics}, pages 189--326. Springer, New York, NY.

\bibitem[Rosenberg et~al., 2007]{rosenberg_using_2007}
Rosenberg, E.~S., Davidian, M., and Banks, H.~T. (2007).
\newblock Using mathematical modeling and control to develop structured treatment interruption strategies for {HIV} infection.
\newblock {\em Drug and Alcohol Dependence}, 88 Suppl 2(Suppl 2):S41--51.

\bibitem[Rosenberger et~al., 2019]{rosenberger2019randomization}
Rosenberger, W.~F., Uschner, D., and Wang, Y. (2019).
\newblock Randomization: {The} forgotten component of the randomized clinical trial.
\newblock {\em Statistics in Medicine}, 38(1):1--12.

\bibitem[Rosthøj et~al., 2006]{rosthoj_estimation_2006}
Rosthøj, S., Fullwood, C., Henderson, R., and Stewart, S. (2006).
\newblock Estimation of optimal dynamic anticoagulation regimes from observational data: a regret-based approach.
\newblock {\em Statistics in Medicine}, 25(24):4197--4215.

\bibitem[Rubin, 1974]{rubin1974estimating}
Rubin, D.~B. (1974).
\newblock Estimating causal effects of treatments in randomized and nonrandomized studies.
\newblock {\em Journal of Educational Psychology}, 66(5):688--701.

\bibitem[Russo et~al., 2018]{russo2017tutorial}
Russo, D.~J., Roy, B.~V., Kazerouni, A., Osband, I., and Wen, Z. (2018).
\newblock A {Tutorial} on {Thompson} {Sampling}.
\newblock {\em Foundations and Trends® in Machine Learning}, 11(1):1--96.

\bibitem[Ryan et~al., 2021]{ryan_defining_2021}
Ryan, J.~C., Viana, J.~N., Sellak, H., Gondalia, S., and O'Callaghan, N. (2021).
\newblock Defining precision health: a scoping review protocol.
\newblock {\em BMJ Open}, 11(2):e044663.

\bibitem[Schulte et~al., 2014]{schulte2014q}
Schulte, P.~J., Tsiatis, A.~A., Laber, E.~B., and Davidian, M. (2014).
\newblock Q- and {A}-learning {Methods} for {Estimating} {Optimal} {Dynamic} {Treatment} {Regimes}.
\newblock {\em Statistical Science: A Review Journal of the Institute of Mathematical Statistics}, 29(4):640--661.

\bibitem[Shortreed et~al., 2014]{shortreed_multiple_2014}
Shortreed, S.~M., Laber, E., Scott~Stroup, T., Pineau, J., and Murphy, S.~A. (2014).
\newblock A multiple imputation strategy for sequential multiple assignment randomized trials.
\newblock {\em Statistics in Medicine}, 33(24):4202--4214.

\bibitem[Strecher et~al., 2008]{strecher_web-based_2008}
Strecher, V.~J., McClure, J.~B., Alexander, G.~L., Chakraborty, B., Nair, V.~N., Konkel, J.~M., Greene, S.~M., Collins, L.~M., Carlier, C.~C., Wiese, C.~J., Little, R.~J., Pomerleau, C.~S., and Pomerleau, O.~F. (2008).
\newblock Web-based smoking-cessation programs: results of a randomized trial.
\newblock {\em American Journal of Preventive Medicine}, 34(5):373--381.

\bibitem[Sugiyama, 2015]{sugiyama2015statistical}
Sugiyama, M. (2015).
\newblock {\em Statistical {Reinforcement} {Learning}: {Modern} {Machine} {Learning} {Approaches}}.
\newblock Chapman and Hall/CRC.

\bibitem[Sutton and Barto, 2018]{sutton18_rl}
Sutton, R.~S. and Barto, A.~G. (2018).
\newblock {\em Reinforcement {Learning}: {An} {Introduction}}.
\newblock Adaptive {Computation} and {Machine} {Learning} series. The MIT Press, Cambridge, Massachusetts, 2 edition.

\bibitem[Tao and Wang, 2017]{tao_adaptive_2017}
Tao, Y. and Wang, L. (2017).
\newblock Adaptive contrast weighted learning for multi-stage multi-treatment decision-making.
\newblock {\em Biometrics}, 73(1):145--155.

\bibitem[Tao et~al., 2018]{tao2018tree}
Tao, Y., Wang, L., and Almirall, D. (2018).
\newblock Tree-based reinforcement learning for estimating optimal dynamic treatment regimes.
\newblock {\em The Annals of Applied Statistics}, 12(3).

\bibitem[Tewari and Murphy, 2017]{tewari2017ads}
Tewari, A. and Murphy, S.~A. (2017).
\newblock From {Ads} to {Interventions}: {Contextual} {Bandits} in {Mobile} {Health}.
\newblock In Rehg, J.~M., Murphy, S.~A., and Kumar, S., editors, {\em Mobile {Health}}, pages 495--517. Springer International Publishing, Cham.

\bibitem[Thall et~al., 2007]{thall2007bayesian}
Thall, P.~F., Wooten, L.~H., Logothetis, C.~J., Millikan, R.~E., and Tannir, N.~M. (2007).
\newblock Bayesian and frequentist two-stage treatment strategies based on sequential failure times subject to interval censoring.
\newblock {\em Statistics in Medicine}, 26(26):4687--4702.

\bibitem[Thompson, 1933]{thompson_likelihood_1933}
Thompson, W.~R. (1933).
\newblock On the {Likelihood} that {One} {Unknown} {Probability} {Exceeds} {Another} in {View} of the {Evidence} of {Two} {Samples}.
\newblock {\em Biometrika}, 25(3/4):285.

\bibitem[Tomczak et~al., 2015]{tomczak_cancer_2015}
Tomczak, K., Czerwińska, P., and Wiznerowicz, M. (2015).
\newblock The {Cancer} {Genome} {Atlas} ({TCGA}): an immeasurable source of knowledge.
\newblock {\em Contemporary Oncology (Poznan, Poland)}, 19(1A):A68--77.

\bibitem[Tsiatis et~al., 2021]{tsiatis_dynamic_2021}
Tsiatis, A.~A., Davidian, M., Holloway, S.~T., and Laber, E.~B. (2021).
\newblock {\em Dynamic {Treatment} {Regimes}: {Statistical} {Methods} for {Precision} {Medicine}}.
\newblock Chapman \& Hall/CRC, Boca Raton.
\newblock OCLC: 1259526921.

\bibitem[Vamathevan et~al., 2019]{vamathevan_applications_2019}
Vamathevan, J., Clark, D., Czodrowski, P., Dunham, I., Ferran, E., Lee, G., Li, B., Madabhushi, A., Shah, P., Spitzer, M., and Zhao, S. (2019).
\newblock Applications of machine learning in drug discovery and development.
\newblock {\em Nature Reviews Drug Discovery}, 18(6):463--477.

\bibitem[van~der Laan and Petersen, 2007]{van_der_laan_statistical_2007}
van~der Laan, M.~J. and Petersen, M.~L. (2007).
\newblock Statistical learning of origin-specific statically optimal individualized treatment rules.
\newblock {\em The International Journal of Biostatistics}, 3(1):Article 6.

\bibitem[van~der Laan et~al., 2007]{van_der_laan_super_2007}
van~der Laan, M.~J., Polley, E.~C., and Hubbard, A.~E. (2007).
\newblock Super learner.
\newblock {\em Statistical Applications in Genetics and Molecular Biology}, 6:Article25.

\bibitem[Voils et~al., 2012]{voils_informing_2012}
Voils, C.~I., Chang, Y., Crandell, J., Leeman, J., Sandelowski, M., and Maciejewski, M.~L. (2012).
\newblock Informing the dosing of interventions in randomized trials.
\newblock {\em Contemporary Clinical Trials}, 33(6):1225--1230.

\bibitem[Wang et~al., 2012]{wang2012evaluation}
Wang, L., Rotnitzky, A., Lin, X., Millikan, R.~E., and Thall, P.~F. (2012).
\newblock Evaluation of {Viable} {Dynamic} {Treatment} {Regimes} in a {Sequentially} {Randomized} {Trial} of {Advanced} {Prostate} {Cancer}.
\newblock {\em Journal of the American Statistical Association}, 107(498):493--508.

\bibitem[Watkins, 1989]{watkins89_rl}
Watkins, C. J. C.~H. (1989).
\newblock {\em Learning from {Delayed} {Rewards}}.
\newblock {PhD}, King's College, Cambridge University, Cambridge, UK.

\bibitem[Yu et~al., 2023]{yu_reinforcement_2023}
Yu, C., Liu, J., Nemati, S., and Yin, G. (2023).
\newblock Reinforcement {Learning} in {Healthcare}: {A} {Survey}.
\newblock {\em ACM Computing Surveys}, 55(1):1--36.

\bibitem[Zhang et~al., 2012]{zhang2012robust}
Zhang, B., Tsiatis, A.~A., Laber, E.~B., and Davidian, M. (2012).
\newblock A {Robust} {Method} for {Estimating} {Optimal} {Treatment} {Regimes}.
\newblock {\em Biometrics}, 68(4):1010--1018.

\bibitem[Zhang et~al., 2013]{zhang2013robust}
Zhang, B., Tsiatis, A.~A., Laber, E.~B., and Davidian, M. (2013).
\newblock Robust estimation of optimal dynamic treatment regimes for sequential treatment decisions.
\newblock {\em Biometrika}, 100(3):681--694.

\bibitem[Zhang et~al., 2020]{zhang_multicategory_2020}
Zhang, C., Chen, J., Fu, H., He, X., Zhao, Y.-Q., and Liu, Y. (2020).
\newblock Multicategory {Outcome} {Weighted} {Margin}-based {Learning} for {Estimating} {Individualized} {Treatment} {Rules}.
\newblock {\em Statistica Sinica}, 30:1857--1879.

\bibitem[Zhao et~al., 2009]{zhao_reinforcement_2009}
Zhao, Y., Kosorok, M.~R., and Zeng, D. (2009).
\newblock Reinforcement learning design for cancer clinical trials.
\newblock {\em Statistics in Medicine}, 28(26):3294--3315.

\bibitem[Zhao et~al., 2012]{zhao2012estimating}
Zhao, Y., Zeng, D., Rush, A.~J., and Kosorok, M.~R. (2012).
\newblock Estimating {Individualized} {Treatment} {Rules} {Using} {Outcome} {Weighted} {Learning}.
\newblock {\em Journal of the American Statistical Association}, 107(499):1106--1118.

\bibitem[Zhao et~al., 2011]{zhao_reinforcement_2011}
Zhao, Y., Zeng, D., Socinski, M.~A., and Kosorok, M.~R. (2011).
\newblock Reinforcement {Learning} {Strategies} for {Clinical} {Trials} in {Non}-small {Cell} {Lung} {Cancer}.
\newblock {\em Biometrics}, 67(4):1422--1433.

\bibitem[Zhao et~al., 2020]{zhao_constructing_2020}
Zhao, Y., Zhu, R., Chen, G., and Zheng, Y. (2020).
\newblock Constructing dynamic treatment regimes with shared parameters for censored data.
\newblock {\em Statistics in Medicine}, 39(9):1250--1263.

\bibitem[Zhao et~al., 2015]{zhao2015new}
Zhao, Y.-Q., Zeng, D., Laber, E.~B., and Kosorok, M.~R. (2015).
\newblock New {Statistical} {Learning} {Methods} for {Estimating} {Optimal} {Dynamic} {Treatment} {Regimes}.
\newblock {\em Journal of the American Statistical Association}, 110(510):583--598.

\bibitem[Zhou et~al., 2018]{zhou2018personalizing}
Zhou, M., Mintz, Y., Fukuoka, Y., Goldberg, K., Flowers, E., Kaminsky, P., Castillejo, A., and Aswani, A. (2018).
\newblock Personalizing {Mobile} {Fitness} {Apps} using {Reinforcement} {Learning}.
\newblock {\em CEUR Workshop Proceedings}, 2068.

\bibitem[Zhou et~al., 2017]{zhou2017residual}
Zhou, X., Mayer-Hamblett, N., Khan, U., and Kosorok, M.~R. (2017).
\newblock Residual {Weighted} {Learning} for {Estimating} {Individualized} {Treatment} {Rules}.
\newblock {\em Journal of the American Statistical Association}, 112(517):169--187.

\end{thebibliography}

\end{document}